\documentclass{article}
\usepackage[a4paper, margin=1in]{geometry}

\newcommand{\smalllab}{\footnotesize}
\newcommand{\xxxx}{\texttt{GraViti}}
\usepackage[inkscapelatex=false]{svg}
\usepackage[english]{babel}
\usepackage{amssymb}
\usepackage{amsmath}
\usepackage{tikz}
\usepackage{multirow}
\usepackage{subcaption}
\usepackage{booktabs}
\usepackage{float}
\usepackage{natbib}

\makeatletter
\newcommand\footnoteref[1]{\protected@xdef\@thefnmark{\ref{#1}}\@footnotemark}
\makeatother

\usetikzlibrary{arrows.meta, positioning, calc, fit, backgrounds}
\usepackage{graphicx}
\usepackage[colorlinks=true, allcolors=blue]{hyperref}

\title{GraViti: Graph-Level Variational Autoencoders\\ with Relaxed Permutation Invariance}
\author{
\textbf{Roman Bresson}\footnote{equal contribution} \footnote{Work started significantly while at KTH Royal Institute of Technology}\\
Mohamed bin Zayed University of Artificial Intelligence
\and
\textbf{Konstantinos Divriotis}$^*$\\
Mohamed bin Zayed University of Artificial Intelligence
\and
\textbf{Johannes F. Lutzeyer}\\
LIX, CNRS, École Polytechnique, IP Paris
\and
\textbf{Iakovos Evdaimon}\\
LIX, CNRS, École Polytechnique, IP Paris
\and
\textbf{Michalis Vazirgiannis}\\
Mohamed bin Zayed University of Artificial Intelligence\\
LIX, CNRS, École Polytechnique, IP Paris
}

\date{\vspace{-5ex}}

\begin{document}
\maketitle
\begin{abstract}
We introduce GraViti, a transformer-based graph-level variational autoencoder that encodes entire graphs into single, fixed-dimensional latent vectors rather than per-node embeddings, yielding a graph-level latent space that supports smooth interpolation, property-guided search, and other downstream tasks beyond the reach of node-level approaches. GraViti achieves state-of-the-art reconstruction accuracy on large molecular graph datasets while offering competitive single-step generative performance. Our central finding concerns where permutation invariance is needed in such a model, and where it is not. The encoder remains permutation-invariant throughout, ensuring the model generalizes to graphs regardless of input order. The reconstruction loss, however, does not need to be: when training data is provided in a consistent node order, comparing predictions to targets directly in that order, without a graph-matching step, is sufficient. Relaxing invariance in the loss alone lets GraViti avoid the cubic-cost matching step required by prior graph-level autoencoders, reducing complexity to quadratic while improving reconstruction fidelity. We show the resulting latent space is chemically meaningful, supporting controlled molecular editing and property optimization, recovering established chemical trends, and enabling direct regression of graph-level properties from latent embeddings.
\end{abstract}

\section{Introduction}

Autoencoders \citep{autoencoders_basic} are unsupervised learning models designed to learn compact representations of complex data. They consist of two components: an encoder that maps an input data point (such as an image, or in our case a graph) into a low‑dimensional latent vector, and a decoder that reconstructs the original data from this latent code. The model is trained end‑to‑end by minimizing the difference between the input and its reconstruction. This training process forces the latent representation to capture the essential structure of the data while discarding redundancy and noise. As a result, autoencoders are widely used for tasks such as denoising, compression, and extracting meaningful features that can support downstream applications \citep{Berahmand2024,LI2023110176}. To enforce structure on the latent representations, variational autoencoders \citep{autoencodingvariational} constrain the distribution of all data embeddings $z$ to match a fixed prior (usually normal) through minimizing a KL-divergence term.

While autoencoders have been applied across various modalities \citep{NEURIPS2024_4f3cb957}, graph autoencoding remains challenging due to graphs' inherently discrete and non-euclidean nature. Most approaches covering graph-oriented AE and VAEs do so at node-level. The early approach by Kipf et al. \citep{kipf_variational_2016} encodes a graph of $N$ nodes into $N$ latent vectors, each encoding one node's information, and the decoder recovers the adjacency matrix. More recent methods still follow this approach \citep{gen3,ketata2025lift,Rigoni2025}, which significantly limits the ability to capture meaningful distances between graphs or manifold structure \citep{sun2019infograph}.

While graph‑level autoencoders (which encode an entire graph in a single latent representation) remain under-explored, with PIGVAE \citep{NEURIPS2021_4f3d7d38} and GRALE \citep{grale} being notable exceptions which will be further discussed in this work. This can be traced to several core challenges inherent to their design. First, the encoder must be permutation‑invariant, since node order is arbitrary while graph structure is not; without this property, the latent vector risks encoding artifacts of node indexing rather than genuine topology. Second, the model must handle variable graph sizes: the encoder must accept graphs of arbitrary cardinality, and the decoder must expand a single latent vector into a graph whose size is not known in advance, a requirement that makes graph‑level decoding substantially more complex than its node‑level counterpart.

In this work, we introduce GraViti, a graph-level VAE that encodes entire graphs into single latent vectors using a graph-transformer backbone. Our contributions are threefold. First, we propose a scalable architecture that achieves state-of-the-art reconstruction accuracy on large graph datasets. Second, we show that the learned latent space is well structured: on molecular benchmarks, it supports interpolation, controlled editing, and property-driven optimization, recovers well-established chemical phenomena, and provides competitive single-step molecular generation. Third, we show through ablations that this improvement stems specifically from relaxing permutation invariance in the reconstruction loss, while keeping the encoder itself permutation-invariant. This relaxation only requires that the training data follows a consistent node ordering, not that the ordering rule itself be known, a property that holds for many real-world graph datasets.

\section{Background and state of the art}

\paragraph{Permutation-invariant encoding of size-varying graphs}

Early GNNs learn permutation equivariant node representations via message passing with permutation-invariant aggregation \citep{kipf2017semi,xu2019how}, including attention-based aggregators that allow neighbor-specific information transmission \citep{NIPS2017_3f5ee243,velikovi2017graph,NEURIPS2020_94aef384}.

Since message-passing GNNs have limited expressivity \citep{xu2019how}, graph transformers were introduced, letting all nodes attend to all others at the cost of a full attention matrix \citep{dwivedi2021generalization,muller2024attending}, with positional encodings now carrying the structural information \citep{NEURIPS2021_b4fd1d2c,huang2025what}. Hybrid approaches such as GraphGPS \citep{10.5555/3600270.3601324} and SAT \citep{Chen22a} combine transformers with message passing to balance efficiency and flexibility. Both families operate directly on sets of nodes and edges, giving permutation-equivariant representations that handle graphs of arbitrary size without a fixed ordering, making them suitable graph encoders as in the original Graph VAE \citep{kipf_variational_2016}. A permutation-invariant pooling operation then produces a graph-level representation, though often at the cost of discarding substantial structural information \citep{10.5555/3327345.3327389}.

\paragraph{Size-varying and permutation-invariant decoding} \label{sec:pigvaegrale}

Graph (variational) autoencoders must reconstruct a graph of unknown size and determine the correspondence between reconstructed and input nodes to compute a reconstruction loss. Node-level approaches handle this naturally: the encoder produces one latent vector per node, and the decoder reconstructs adjacency and attributes in the same fixed order, so correspondence is implicit in node indexing. Common decoders include inner-product or MLP heads \citep{kipf_variational_2016}, sometimes refined via message passing \citep{pmlr-v97-grover19a}.

Graph-level autoencoders lack this implicit ordering and must instead resolve it explicitly. Graph matching \citep{Simonovsky2018,huang2026towards} finds an optimal node alignment, at the cost of solving an expensive quadratic assignment problem. PIGVAE \citep{NEURIPS2021_4f3d7d38} instead learns a relaxed permutation matrix by sorting node embeddings, avoiding the assignment problem but offering only limited alignment; GRALE \citep{grale} improves on this with a Sinkhorn-based matcher and an optimal-transport reconstruction loss, yielding a full doubly stochastic alignment matrix with better stability and expressivity.

Variable graph size is typically handled by padding to a fixed maximum, as in NLP transformers \citep{NIPS2017_3f5ee243}, often via a ``dummy node" class \citep{grale,9516695}, with the decoder predicting a maximal-size graph plus a node mask. This has two drawbacks: it prevents generalization to larger graphs, and the dummy class biases node classification with an artificial label. PIGVAE instead broadcasts the graph representation across a sequence transformer, using sinusoidal positional encodings to differentiate duplicated tokens; this avoids both a maximum size and a dummy class, masking only non-existent nodes, with size predicted by an MLP from the latent.

Overall, existing methods trade off scalability against expressivity: node-level approaches scale well but yield incoherent graph-level latent spaces, while graph-level methods need a costly matching step. This motivates methods that learn meaningful graph-level representations without expensive alignment, while remaining expressive and scalable.

\section{The \xxxx~architecture}

\paragraph{Graph representation}

Let $N$ denote the size of the graph (i.e., the number of nodes). In this work, we assume that graphs have discrete node classes and edge classes instead of continuous features (e.g. atom classes, color). A graph is represented by two objects. First is its node-class matrix $X \in \mathbb{R}^{N \times n_c}$ where $n_c$ is the number of node classes and $X_i$ is the one-hot encoding of the $i$-th node's class. Importantly, we do not introduce a ``dummy node" class; all classes correspond to real nodes. Then $E \in \mathbb{R}^{N \times N \times n_e}$ is the edge class tensor, where $n_e$ is the number of edge classes, including both actual edge types and an additional class representing ``no edge".

To enrich the graph representation, we compute directly from the graph a global feature vector $y \in \mathbb{R}^{d_y}$ containing graph-level information. In practice, $d_y = n_c + n_e + 1$, and $y$ encodes the count of each node and edge type, as well as the graph size. We also compute $P \in \mathbb{R}^{N \times d_P}$, the positional encodings for the nodes, obtained by concatenating Laplacian positional encodings and random walk positional encodings. These inject global structural information into the graph transformer at no extra cost.

Finally, when relevant for the field, we also exploit a matrix of node-specific features derived from domain knowledge $K \in \mathbb{R}^{N \times d_k}$. For molecular datasets, these include atomic properties such as atomic number, atomic mass, valence electrons, and electronegativity.

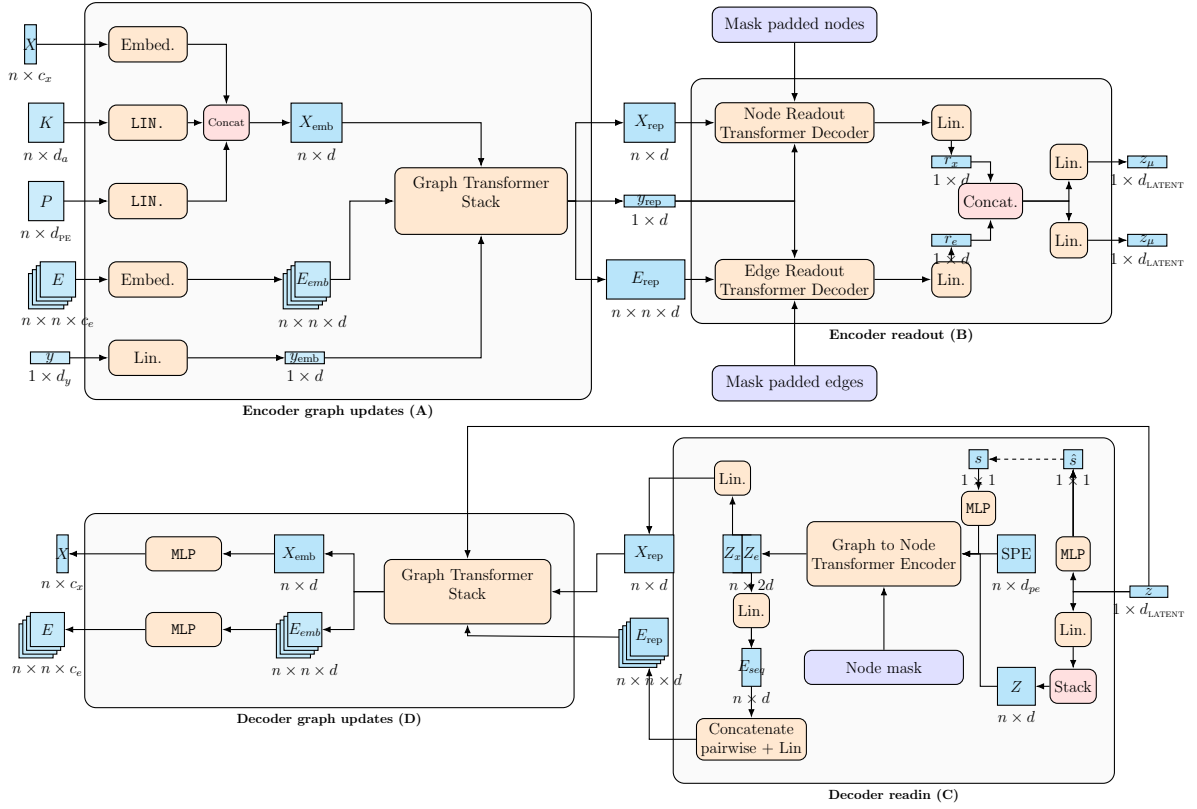
\begin{figure}[t]
    \centering
\resizebox{\textwidth}{!}{
\tikzset{
  >={Latex[length=2.0mm, width=1.6mm]},
  line/.style={draw, -{Latex}, thick},
  box/.style={draw, rounded corners=2mm, thick, align=center, minimum height=9mm, minimum width=22mm},
  group/.style={draw, rounded corners=3mm, thick, inner sep=6mm, fill=black!2},
  trainable/.style={box, fill=orange!18},
  nontrainable/.style={box, fill=red!12},
  quantity/.style={box, fill=blue!12, minimum width=12mm},
  smalllab/.style={font=\footnotesize},
  titlelab/.style={font=\bfseries},
}

\newcommand{\tensorstack}[9]{%
  \begin{scope}
    \coordinate (#1-sw) at (#2+#4/2,#3-#5/2);
    \pgfmathtruncatemacro{\L}{#6-1}
    \coordinate (#1-east) at (#2+#4+#4/2+0.1*\L,#3+0.1*\L);
    \coordinate (#1-west) at (#2+#4/2,#3+0.1*\L);
    \coordinate (#1-north) at (#2+#4,#3+0.1*\L+#5/2);
    \coordinate (#1-south) at (#2+#4+0.1*\L,#3-#5/2);
    \foreach \i in {0,...,\L}{
      \draw[thick, fill=#8]
        ($(#1-sw)+(\i*0.1,\i*0.1)$) rectangle ++(#4,#5);
    }
    \coordinate (#1) at ($(#1-sw)+(\L*0.1,\L*0.1)+(#4/2,#5/2)$);
    \node at (#1) {#9}; 
      \node[below=1pt of #1-south] {#7}; 
  \end{scope}
}

\begin{tikzpicture}[font=\large]


\tensorstack{Xin}{0.2}{4}{0.3}{1.0}{1}{$n\times c_x$}{cyan!25}{$X$}

\tensorstack{Ein}{0.}{-2.3}{0.9}{0.9}{4}{$n\times n\times c_e$}{cyan!20}{$E$}

\tensorstack{Kin}{0}{2}{0.9}{1.0}{1}{$n\times d_a$}{cyan!20}{$K$}

\tensorstack{Pin}{0.0}{0}{0.9}{1.0}{1}{$n\times d_{\textsc{pe}}$}{cyan!18}{$P$}

\tensorstack{yin}{0.0}{-4}{1.0}{0.3}{1}{$1\times d_y$}{cyan!18}{$y$}

\node[trainable, minimum width=20mm] (xemb) at (3.5,4)
{Embed.};
\draw[line] (Xin-east) -- (xemb.west);

\node[trainable, minimum width=20mm] (kmlp) at (3.5,2)
{\texttt{LIN.}};
\draw[line] (Kin-east) -- (kmlp.west);

\node[trainable, minimum width=20mm] (pmlp) at (3.5,0)
{\texttt{LIN.}};
\draw[line] (Pin-east) -- (pmlp.west);

\node[nontrainable, minimum width=9mm] (concatX) at (5.5,2.)
{{\smalllab Concat}};
\draw[line] (xemb.east) -| (concatX.north);
\draw[line] (kmlp.east) -- (concatX.west);
\draw[line] (pmlp.east) -| (concatX.south);

\tensorstack{Xemb}{6.5}{2}{1.3}{1.0}{1}{$n\times d$}{cyan!25}{$X_{\mathrm{emb}}$}
\draw[line] (concatX.east) -- (Xemb-west);

\node[trainable, minimum width=20mm] (eemb) at (3.5,-2)
{Embed.};
\draw[line] (Ein-east) -- (eemb.west);

\tensorstack{Eemb}{6.5}{-2.3}{0.9}{0.9}{4}{$n\times n\times d$}{cyan!20}{$E_{emb}$}
\draw[line] (eemb.east) -- (Eemb-west);
%
\node[trainable, minimum width=20mm] (yproj) at (3.5,-4)
{Lin.};
\draw[line] (yin-east) -- (yproj.west);

\tensorstack{yemb}{6.5}{-4}{1.0}{0.3}{1}{$1\times d$}{cyan!18}{$y_{\mathrm{emb}}$};
\draw[line] (yproj.east) -- (yemb-west);


\node[trainable, minimum width=44mm, minimum height=17mm] (GT)
at (12,0)
{Graph Transformer\\Stack\\};

\draw[line] (Xemb-east) -| (GT.north);
\draw[line] (Eemb-east) -| ++ (5mm,0) |- (GT.west);
\draw[line] (yemb-east) -| (GT.south);

\tensorstack{Xrep}{15}{2}{1.3}{1.0}{1}{$n\times d$}{cyan!25}{$X_{\mathrm{rep}}$}

\tensorstack{Erep}{14.2}{-2}{2.}{1.0}{1}{$n\times n\times d$}{cyan!25}{$E_{\mathrm{rep}}$}

\tensorstack{yrep}{15}{0}{1.3}{0.3}{1}{$1\times d$}{cyan!25}{$y_{\mathrm{rep}}$}

\draw[line] (GT.east) -- ++(2mm,0) |- (Xrep-west);
\draw[line] (GT.east) -- ++(2mm,0) |- (Erep-west);
\draw[line] (GT.east) -- ++(2mm,0) -- (yrep-west);

\begin{scope}[on background layer]
  \node[group,
        fit=(xemb)(kmlp)(pmlp)(concatX)(Xemb)
            (eemb)(Eemb)
            (yproj)(yemb)
            (GT), label={[titlelab]south:Encoder graph updates (A)}] (trunkbg) {} ;
\end{scope}


\node[trainable, minimum width=40mm] (nodeRO) at (20,2)
{Node Readout\\Transformer Decoder};

\node[trainable, minimum width=40mm] (edgeRO) at (20,-2.)
{Edge Readout\\Transformer Decoder};

\draw[line] (Xrep-east) |- (nodeRO.west);
\draw[line] (Erep-east) |- (edgeRO.west);
\draw[line] (yrep-east) -| ++(4mm,0) -| (edgeRO.north);
\draw[line] (yrep-east) -| ++(4mm,0) -| (nodeRO.south);

\node[trainable, minimum width=10mm] (normprojX) at (24, 2) {Lin.};
\node[trainable, minimum width=10mm] (normprojE) at (24,-2) {Lin.};

\draw[line] (nodeRO.east) -- (normprojX.west);
\draw[line] (edgeRO.east) -- (normprojE.west);

\tensorstack{rX}{23}{1}{1.0}{0.3}{1}{$1 \times d$}{cyan!25}{$r_x$};
\tensorstack{rE}{23}{-1}{1.0}{0.3}{1}{$1 \times d$}{cyan!25}{$r_e$};

\draw[line] (normprojX.south) -- (rX-north);
\draw[line] (normprojE.north) -- (rE-south);

\node[nontrainable, minimum width=10mm] (fuse) at (25,0) {Concat.};
\draw[line] (rX-east) -- ++(1mm,0) -| (fuse.north);
\draw[line] (rE-east) -- ++(1mm,0) -| (fuse.south);

\node[trainable, minimum width=9mm] (headmu) at (27,1) {Lin.};
\node[trainable, minimum width=9mm] (headlv) at (27,-1) {Lin.};

\tensorstack{zmu}{28}{1}{1.0}{0.3}{1}{$1 \times d_{\textsc{latent}}$}{cyan!25}{$z_\mu$}
\tensorstack{zlv}{28}{-1}{1.0}{0.3}{1}{$1 \times d_{\textsc{latent}}$}{cyan!25}{$z_\sigma$}

\draw[line] (fuse.east) -- ++(7mm,0) -| (headmu.south);
\draw[line] (fuse.east) -- ++(7mm,0) -| (headlv.north);
\draw[line] (headmu.east) -- (zmu-west);
\draw[line] (headlv.east) -- (zlv-west);

\begin{scope}[on background layer]
  \node[group,
        fit=(nodeRO)(edgeRO)(normprojX)(normprojE)(rX)(rE)(fuse)(headmu)(headlv),
      label={[titlelab]south:Encoder readout (B)}] (readgrp) {};
\end{scope}
\node[quantity, minimum width=42mm] (masknote) at (20,-4.65)
{Mask padded edges};
\draw[line] (masknote.north) -- (edgeRO.south);

\node[quantity, minimum width=42mm] (masknotenode) at (20,4.5)
{Mask padded nodes};
\draw[line] (masknotenode.south) -- (nodeRO.north);

\end{tikzpicture}}

\resizebox{\textwidth}{!}{
\tikzset{
  >={Latex[length=2.0mm, width=1.6mm]},
  line/.style={draw, {Latex}- , thick},
  box/.style={draw, rounded corners=2mm, thick, align=center, minimum height=9mm, minimum width=22mm},
  group/.style={draw, rounded corners=3mm, thick, inner sep=6mm, fill=black!2},
  trainable/.style={box, fill=orange!18},
  nontrainable/.style={box, fill=red!12},
  quantity/.style={box, fill=blue!12, minimum width=12mm},
  smalllab/.style={font=\footnotesize},
  titlelab/.style={font=\bfseries},
}

\newcommand{\tensorstack}[9]{%
  \begin{scope}
    \coordinate (#1-sw) at (#2+#4/2,#3-#5/2);
    \pgfmathtruncatemacro{\L}{#6-1}
    \coordinate (#1-east) at (#2+#4+#4/2+0.1*\L,#3+0.1*\L);
    \coordinate (#1-west) at (#2+#4/2,#3+0.1*\L);
    \coordinate (#1-north) at (#2+#4,#3+0.1*\L+#5/2);
    \coordinate (#1-south) at (#2+#4+0.1*\L,#3-#5/2);
    \foreach \i in {0,...,\L}{
      \draw[thick, fill=#8]
        ($(#1-sw)+(\i*0.1,\i*0.1)$) rectangle ++(#4,#5);
    }
    \coordinate (#1) at ($(#1-sw)+(\L*0.1,\L*0.1)+(#4/2,#5/2)$);
    \node at (#1) {#9}; 
      \node[below=1pt of #1-south] {#7}; 
  \end{scope}
}

\newcommand{\qtensor}[9]{%
  \tensorstack{#1}{#2}{#3}{#4}{#5}{#6}{0.1}{#8}{#9}%
  \coordinate (#1-west)  at (#2,#3);
  \coordinate (#1-east)  at (#2 + #4 + #4/2 + 0.1*(#6-1), #3 + 0.1*(#6-1));
  \coordinate (#1-north) at (#2 + #4/2 + 0.1*(#6-1), #3 + #5/2 + 0.1*(#6-1));
  \coordinate (#1-south) at (#2 + #4/2 + 0.1*(#6-1), #3 - #5/2 + 0.1*(#6-1));
}

\begin{tikzpicture}[font=\large]


\tensorstack{Xin}{1}{1}{0.3}{1.0}{1}{$n\times c_x$}{cyan!25}{$X$}

\tensorstack{Ein}{-0.3}{-1.3}{0.9}{0.9}{4}{$n\times n\times c_e$}{cyan!20}{$E$}

\node[trainable, minimum width=20mm] (xclas) at (4.5,1)
{\texttt{MLP}};
\draw[line] (Xin-east) -- (xclas.west);

\tensorstack{Xemb}{6.25}{1}{1.3}{1.0}{1}{$n\times d$}{cyan!25}{$X_{\mathrm{emb}}$}
\draw[line] (xclas.east) -- (Xemb-west);
\node[trainable, minimum width=20mm] (eclas) at (4.5,-1)
{\texttt{MLP}};

\tensorstack{Eemb}{6.5}{-1.3}{0.9}{0.9}{4}{$n\times n \times d$}{cyan!20}{$E_{emb}$}
\draw[line] (Eemb-east) -| ++(8.5mm,0mm) |- (GT.west);
\draw[line] (eclas.east) -- (Eemb-west);
\draw[line] (Xemb-east) -| ++(8mm,0mm) |- (GT.west);
\draw[line] (Ein-east) -- (eclas.west);

\node[trainable, minimum width=44mm, minimum height=17mm] (GT)
at (12,0)
{Graph Transformer\\Stack\\[-1mm]};

\tensorstack{Xrep}{15.5}{1}{1.3}{1.0}{1}{$n\times d$}{cyan!25}{$X_{\mathrm{rep}}$}
\tensorstack{Erep}{15.5}{-1.5}{1.}{1.0}{4}{$n\times n \times d$}{cyan!25}{$E_{\mathrm{rep}}$}

\begin{scope}[on background layer]
  \node[group,
        fit=(Xemb)
            (eemb)(Eemb)
            (GT)(xclas)(eclas), label={[titlelab]south:Decoder graph updates (D)}] (trunkbg) {} ;
\end{scope}

\node[trainable, minimum width=9mm] (Linx) at (19.,3)
{Lin.};

\tensorstack{z}{29}{0}{1.0}{0.3}{1}{$1\times d_{\textsc{latent}}$}{cyan!25}{$z$}
\draw[line] (Xrep-north) -- ++(0mm,1mm) |- (Linx.west);
\draw[line] (GT.east) -- ++(12mm,0) |- (Xrep-west);
\draw[line] (GT.south) |- ++(0mm,-3mm) -- (Erep-west);
\draw[line] (GT.north) |- ++(0mm,35mm) -| (z-north);

\node[trainable, minimum width=9mm] (Line) at (19.5,-0.5)
{Lin.};

\node[trainable, minimum width=40mm, minimum height=17mm] (GTEM) at (23,1)
{Graph to Node\\Transformer Encoder};

\tensorstack{Zstacked_e}{19}{1}{0.5}{1}{1}{$n\times 2d$}{cyan!25}{$Z_e$}
\tensorstack{Zstacked_x}{18.5}{1}{0.5}{1}{1}{}{cyan!25}{$Z_x$}
\draw[line] (Linx.south) |- (Zstacked_e-north) ;
\draw[line] (Line.north) |- (Zstacked_x-south) ;

\node[trainable, minimum width=9mm] (size_pred) at (28,1) {\texttt{MLP}};
\node[trainable, minimum width=9mm] (size_reinj_lin) at (25.5,2.2) {\texttt{MLP}};
\node[trainable, minimum width=9mm] (linstack) at (28,-1) {Lin.};
\tensorstack{eseq}{19}{-2.}{0.5}{1}{1}{$n\times d$}{cyan!25}{$E_{seq}$}
\node[nontrainable, minimum width=9mm] (stack) at (28,-2.5) {Stack};
\tensorstack{Zstacked}{25.5}{-2.5}{1.0}{1.0}{1}{$n\times d$}{cyan!25}{$Z$}
\tensorstack{PosEnc}{25.5}{1}{1.0}{1.0}{1}{$n\times d_{pe}$}{cyan!25}{SPE}
\node[trainable, minimum width=9mm] (LinCat) at (19.5,-3.9) {Concatenate \\ pairwise + Lin};

\tensorstack{size}{27.5}{3.5}{0.5}{0.5}{1}{$1 \times 1$}{cyan!25}{$\hat{s}$}
\draw[line] (eseq-north) -- (Line.south);
\tensorstack{size_reinj}{25.0}{3.5}{0.5}{0.5}{1}{$1 \times 1$}{cyan!25}{$s$}
\draw[line] (eseq-north) -- (Line.south);
\draw[line] (LinCat.north) -- (eseq-south);
\draw[line] (Erep-south) |- ++(0,-1.9cm) -- (LinCat.west);
\draw[line,dashed] (size_reinj-east) -- (size-west);
\draw[line] (size_reinj_lin.north) |- (size_reinj-south);
\draw[line] (GTEM.east) -| (size_reinj_lin.south);
\draw[line] (GTEM.east) -- (PosEnc-west);
\draw[line] (GTEM.east) -| ++(5mm,0) |- (Zstacked-west);
\draw[line] (Zstacked_e-east) -- (GTEM.west);

\node[quantity, minimum width=42mm] (masknotenode) at (23,-2)
{Node mask};
\draw[line] (GTEM.south) -- (masknotenode.north);

\draw[line] (size-south) --(size_pred.north);
\draw[line] (size-south) --(size_pred.north);

\draw[line] (size_pred.south) |- (z-west);
\draw[line] (stack.north) -- (linstack.south);
\draw[line] (linstack.north) |- (z-west);
\draw[line] (Zstacked-east) -- (stack.west);
\draw[line] (size-south) --(size_pred.north);

\begin{scope}[on background layer]
  \node[group,
        fit=(LinCat)(Linx)(Line)(linstack)(rX)(rE)(fuse)(headmu)(headlv),
      label={[titlelab]south:Decoder readin (C)}] (readgrp) {};
\end{scope}

\end{tikzpicture}}
\caption{The architecture of our model. The data space is on the left, while the latent space is on the right. Encoder on top reads left to right, decoder reads right to left.}
\label{fig:modelarchitecture}
\end{figure}

Since training involves batching, all graphs in a batch must share the same size. To achieve this, we pad all graphs to reach the size of the largest graph in the batch, extending $X$, $P$, $K$, and $E$ accordingly. This padding has no semantic meaning, and the model is designed to ignore it entirely. The padded size of the input does not influence the decoder or model size in any way. The model is illustrated in figure \ref{fig:modelarchitecture}; the letters in paragraph titles refer to the respective components on the figure.

\paragraph{Encoding node, edge, and global representation (A)}

Our encoder maps a triplet $X,E,y$ (optionally $P$ and $K$) to a mean and standard deviation $z_\mu,z_\sigma\in \mathbb{R}^{d_z}$ where $d_z$ is the latent dimension. At first, the node representation matrix is expanded into $X_{\text{emb}}$ by concatenating a learned shallow embedding of the node class, both positional encoding (Laplacian and random-walk) processed through a dedicated linear layer and optionally node attributes processed through a dedicated linear layer. Edge representations are obtained via a learned embedding layer of their discrete classes, producing $E_{\text{emb}}$. The global feature vector $y$ is projected through a linear layer into $y_{\text{emb}}$. 

All three tensors ($X_{\text{emb}}$, $E_{\text{emb}}$, $y_{\text{emb}}$) undergo layer normalization and are updated through multiple graph transformer layers, yielding $X_{\text{rep}}$, $E_{\text{rep}}$, and $y_{\text{rep}}$.  This step uses the specific graph transformer architecture introduced in DiGress \citep{conficlrVignacKSWCF23} and updated in CatFlow \citep{NEURIPS2024_15b78035}. This design enables simultaneous updates of node, edge, and graph-level features while ensuring permutation-equivariance.

\paragraph{Encoder readout (B)}

Traditional readout functions for graph-level representations (such as sum, mean, or max pooling) are permutation-invariant at the cost of a large quantity of information. To ensure a highly expressive latent space, we adopt a sequence transformer-based readout mechanism. Design motivations as to why we maintain a permutation-invariant encoder are detailed in Appendix \ref{sec:designchoices}.

Unlike GRALE, which applies a transformer solely on node embeddings, our approach incorporates nodes, edges, and global features into the readout to prevent overloading node embeddings. Specifically, the readout consists of two parallel sequence-decoder transformers: one for nodes and one for edges. Both transformers ignore padded nodes and edges. $y_{\text{rep}}$ serves as the target for both transformers, injecting global context into the readout.

Since positional encodings are omitted at this stage, the inherent permutation invariance of the sequence transformer guarantees that the output is independent of node ordering. Finally, node and edge outputs are normalized independently and projected through linear layers to produce the graph-level representation. This one can be either a single vector (classic autoencoder case) or a pair of same-sized vectors (in the variational case).

\paragraph{Graph size prediction (C)} One crucial information required to reconstruct the graph is its size, which needs to be recovered from the latent representation. We make the design choice of disentangling this crucial structural information from the rest of the latent representation by forcing the model to encode it in only the first $d_s$ dimensions of the latent space. The log-size $\hat{s}$ is then predicted by an MLP. In order for the decoding process to leverage the size information, we use another MLP to re-encode the log-size $s$ into $d_s$ dimensions that replace the corresponding latent features. During training, no gradient flows between $\hat{s}$ and $s$. The predicted log-size $\hat{s}$ is optimized via a smooth L1 loss against the ground-truth log-size, while the ground-truth value $s$ is directly reinjected into the latent representation. This ensures that the corresponding latent dimensions encode size information exclusively, rather than relying on indirect supervision through masking.

\paragraph{Decoding graph-level into node-level representation (C)} The first module of our decoder has to invert the readout function, i.e. obtain node representations from graph representations. Similarly to PIGVAE we first broadcast the graph-level information into a node-wise representation. This is done by simply duplicating our graph-level representation (i.e. the latent representation projected through a linear layer) $N$ times. We therefore obtain a sequence of $N$ identical vectors of dimension $D$. In order to break the symmetries and allow the transformer to differentiate between these pseudo-tokens, we add sinusoidal positional encoding to the sequence. During training, $N$ is determined as the ground truth size, which is also used for reinjection. During inference, $N$ is $\texttt{round}(\exp(\hat{s}))$.

This sequence is then updated into another $N\times D$ sequence through a sequence transformer encoder, matching our readout. We then split this sequence width-wise into two sequences of size $N\times \frac{D}{2}$, a node-oriented sequence $X_x$ and an edge-oriented sequence $X_e$. We create an edge-feature matrix $\phi_e$ by concatenating the rows of $X_e$ in pairs:
$\phi_{e,(i,j)} = X_{e,i} | X_{e,j}$; finally, symmetric edge-feature $\Phi_e$ matrix is obtained as $\Phi_{e,(i,j)} = \frac{1}{2}\left(\phi_{e,(i,j)} + \phi_{e,(j,i)}\right) \in \mathbb{R}^{n\times n \times D}$. In parallel, we create a node features matrix~$\phi_x \in \mathbb{R}^{n\times D}$.

\paragraph{Updating and refining the graph (D)} We update the obtained graph through another graph transformer, with similar architecture as that of the encoder. Finally, we apply an MLP to the nodes and one to the edges as a final prediction head. In our case, those are classifiers. We only reconstruct $X$ and $E$, which is all that is needed for a graph. Also note that the size is predicted, and there is therefore no need to predict an extra dummy node class. The model is trained following the standard VAE objective, {using a cross-entropy loss for all node and edge classes predictions.} A KL term regularizes the latent variables toward a normal prior. Details of the loss formulation are provided in Appendix~\ref{sec:loss}.

\section{Comparison to GRALE and PIGVAE}

\textbf{Loss function} Both GRALE and PIGVAE leverage a permutation-invariant scheme for their loss functions, allowing to reconstruct the graph in any order while still obtaining a relevant reconstruction loss: PIGVAE predicts a permutation matrix obtained through a solver, while GRALE iterates upon this by using a Sinkhorn matcher to obtain a relaxed permutation matrix, and using the Any2Graph loss \citep{krzakala2024endtoend} on the soft assignment. While these approaches are elegant and well-suited to fully general graph settings, this incurs a significant computational overhead (at least cubic for both approaches). In contrast, our model relies on a simpler reconstruction objective made possible by the use of a consistent node ordering, allowing a quadratic complexity. In practice, this results improved reconstruction quality on the benchmarks we consider, especially as graph size increases.

\textbf{Graph size flexibility}
GRALE decodes graphs by iteratively refining a learned node and edge representation of fixed maximal size. This mechanism inherently enforces an upper bound on the size of generated graphs. In contrast, our approach, like PIGVAE, does not impose such a limit: the graph-level latent is replicated to the desired number of nodes, and continuous sequence-based positional encodings are used to distinguish node-level embeddings. As a result, the model architecture is independent of graph size, although computational cost for full attention naturally scales with it.
    
\textbf{Variational component}
Similar to PIGVAE, our method follows a fully variational formulation, enabling likelihood-based training and explicit regularization of the latent space. Unlike GRALE, which relies on a tokenized latent representation, we operate in a continuous latent space. In practice, this enables a well-structured latent, unlocking in particular generative and editing capabilities.

\section{Experimental results} \label{sec:exp}

\paragraph{Experimental setting} A prominent application domain for graph-level representation learning is computational chemistry, where molecules are naturally represented as attributed graphs of atoms and bonds. We consider two benchmark datasets: QM9 \citep{qm9}, a dataset of approximately $130{,}000$ molecules containing up to 9 heavy atoms, and PubChem \citep{PubChem}. Following GRALE, we focus on the PubChem16 and PubChem32 subsets, comprising approximately $13$ million and $74$ million molecules with up to 16 and 32 heavy atoms, respectively. Both datasets are publicly provided in a canonical ordering established by proprietary softwares, as detailed in Appendix \ref{app:datasets} alongside the dataset statistics. As is standard in molecular graph modeling, we remove explicit hydrogen atoms \citep{hydro1:,hydro2,NEURIPS2024_15b78035,hydro3} and { study both the aromatic form and the kekulized form. The resulting graphs contain three bond types (resp. four) in kekulized form (resp. aromatic): single, double, triple (resp. and aromatic). An additional edge type is introduced to denote ``no-edge". We also consider non-molecular graphs through the Bayesian Networks-Asia dataset, which consists in directed acyclic graphs.} Training settings can be found in Appendix \ref{app:training}.

\paragraph{Reconstruction performance}

Table \ref{tab:recon_perf_kek} reports the graph reconstruction performance of our model using what we consider to be the primary evaluation metric of this study. We deliberately adopt a particularly strict reconstruction criterion, under which a graph is deemed correctly reconstructed only if all node types and all edge types are predicted without error. For molecules in aromatic form, this requirement is further extended to include the exact reconstruction of the hydrogen count and formal charge for every aromatic atom. The reported accuracy therefore corresponds to the proportion of graphs that are perfectly reconstructed. We also report the less strict edit distance \citep{JMLR:v25:23-0572} for each model. All models are trained until convergence.

{ We observe that our model outperforms the matching-based baselines on all molecular datasets. We also notice that, as the size of the graphs in the datasets increases, so does the performance gap, with our model beating the baselines by $25\%$ on PubChem32. We attribute this success to the increasing difficulty of training a proper matching as graph size increases for GRALE. This demonstrates in particular that the canonical ordering provided in molecular datasets is sufficiently consistent to ensure good performance, especially as graphs get larger, rejoining results observed on readout functions \citep{buterez2022graph}. Results in aromatic form are similar (see Appendix \ref{app:arom}}). The ASIA-BN dataset was perfectly fitted by all models.

\begin{table}[t]
\caption{Percentage of perfectly reconstructed graphs and edit distance. Molecules are in kekulized form. *PIGVAE's results from \citep{grale}.}




\label{tab:recon_perf_kek}
\centering
{\begin{tabular}{l c c c c c}
\toprule
\textsc{Dataset} & \textsc{Metric} & \textsc{GRALE} & \textsc{PIGVAE}$^*$ & \textsc{GraViti (AE)} & \textsc{GraViti (VAE)}
 \\
\midrule
\multirow{2}{*}{QM9NoHydro}
  & Edit Dist. & \textbf{0.08} & N/A & 0.11 & 0.10\\
  & Accuracy   & {95.7} & N/A & 97.6 & \textbf{97.8}\\
\midrule
\multirow{2}{*}{PubChem16}
  & Edit Dist. & 0.06 & 1.69$^*$ & \textbf{0.04} & \textbf{0.04} \\
  & Accuracy   & 96.9 & 41.0$^*$ & \textbf{99.5} & \textbf{99.5} \\
\midrule
\multirow{2}{*}{PubChem32}
  & Edit Dist. & 0.70 & 2.53$^*$ & 0.15 & \textbf{0.14} \\
  & Accuracy   &  73.3 & 24.9$^*$ & 98.2 & \textbf{98.4} \\
\midrule
\multirow{2}{*}{ASIA-BN}
  & Edit Dist. & \textbf{0}  & N/A & 0 & 0  \\
  & Accuracy   & \textbf{100}  & N/A & 100   & 100 \\
\bottomrule 
\end{tabular}}
\end{table}

\paragraph{Performance and efficiency}

\begin{table}[t]
\centering
\caption{Training efficiency comparison between GRALE and GraViti on PubChem.}
\label{tab:time_comparison}
\begin{tabular}{llcccc}
\toprule
Dataset & Model & Params & Epochs & Accuracy (\%) & Time (h) \\
\midrule
\multirow{2}{*}{PubChem16}
& GRALE   & 7M  & 5 & 94.6 & 44.0 \\
& GraViti & 7M  & 5 & 98.4 & 37.5 \\
\midrule
\multirow{2}{*}{PubChem32}
& GRALE   & 35M & 5 & 73.3 & 229.8 \\
& GraViti & 35M & 5 & 97.5 & 186.5 \\
\bottomrule
\end{tabular}
\end{table}

On top of the higher performance ceiling, we further demonstrate the efficiency gains obtained by leveraging a canonical ordering. To this end, we train two models of comparable size, one GRALE and one GraViti, for the same number of epochs. The results, reported in Table~\ref{tab:time_comparison}, show that eliminating the matching step leads to a substantial reduction in training time while simultaneously yielding higher reconstruction accuracy. These gains become even more pronounced as graph size increases.

\paragraph{Ablation studies}

{
To determine whether our gains stem from architectural choices or from the canonical ordering used in the loss, we ablate several components of the architecture (Table~\ref{tab:ablations}). The model proves remarkably robust to most modifications. The largest drop comes from replacing the transformer encoder with a message-passing GNN, though this variant still outperforms GRALE. Training with GRALE's matching-based loss instead, however, yields 96.5\% accuracy, matching GRALE's 96.9\%, which strongly suggests that the reconstruction objective accounts for most of the improvement.
}

\begin{table}[t]
\centering
\caption{Architectural ablation study results. The baseline model is the VAE on PubChem16}
\label{tab:ablations}
\begin{tabular}{l l c c c c}
\toprule
\textsc{Ablated component} &
\textsc{Alternative} &
\textsc{Acc.} &
\textsc{Diff.} &
\textsc{Ed. Dist.} &
\textsc{Diff.} \\
\midrule
Baseline  & & 99.5 &  0  & 0.04 & 0  \\
Transformer Encoder & GNN-based Enc. & 97.2 &  -2.3 & 0.15  & +0.1 \\
Transformer Decoder & MLP Decoder & 99.3 & -0.2 & 0.06 & +0.01 \\
Atom attributes Used & Not used & 99.4 & -0.1 & 0.05 & +0.01\\
Laplacian + RW Pos. Enc. & Laplacian only & 99.2 & -0.3 & 0.05 & +0.01 \\
Laplacian + RW Pos. Enc. & RW only & 99.5 & 0 & 0.04 & 0 \\
Laplacian + RW Pos. Enc. & None & 99.2 & -0.3 & 0.06 & +0.02 \\
Global features & No global features & 99.4 & -0.1 & 0.05 & +0.01\\
\bottomrule
\end{tabular}
\end{table}

\paragraph{Limitations of Canonical Ordering Dependance}

{
To illustrate the limits of our scope, we test on COLORING, a very general dataset with no natural node ordering, where our model should struggle. As expected, baseline reconstruction accuracy is near $0\%$. Imposing a generic, domain-agnostic WL-color-based node ordering (details in Appendix~\ref{sec:orderingcolor}) raises accuracy to $80.5\%$, while remaining significantly below matching-based models. Table \ref{tab:recon_color} reports the full results alongside comparisons to matching-based baselines. This shows a middle ground between no ordering at all, where our model fails outright, and the well-informed, domain-specific canonical orderings available for molecules, where our model achieves state-of-the-art performance.

\begin{table}[t]
\caption{Percentage of perfectly reconstructed graphs and edit distance for the Coloring dataset. Results for PIGVAE are reported from \citep{grale}.}

\label{tab:recon_color}
\centering
{\begin{tabular}{l c c c c c}
\toprule
\textsc{Dataset} & \textsc{Metric} & \textsc{GRALE} & \textsc{PIGVAE}$^*$ & \textsc{Ours (AE)} & \textsc{Ours (VAE)}
 \\
\midrule
\multirow{2}{*}{Coloring no order.}
  & Edit Dist. & \textbf{0.11}  & 2.13$^*$ & 35.5 & 35.3  \\
  & Accuracy   & \textbf{95.8}  & 85.3$^*$ & 0    & 0 \\
\midrule
\multirow{2}{*}{Coloring + order.}
  & Edit Dist. & \textbf{0.06}  & N/A & 0.57 & 0.79 \\
  & Accuracy   & \textbf{96.9}  & N/A & 80.5 & 76.3 \\
\bottomrule 
\end{tabular}}
\end{table}}

\paragraph{Interpolation}

\begin{figure}[t]
    \centering
    \begin{minipage}[c]{0.48\linewidth}
        \centering
        \fbox{\includegraphics[width=0.9\linewidth]{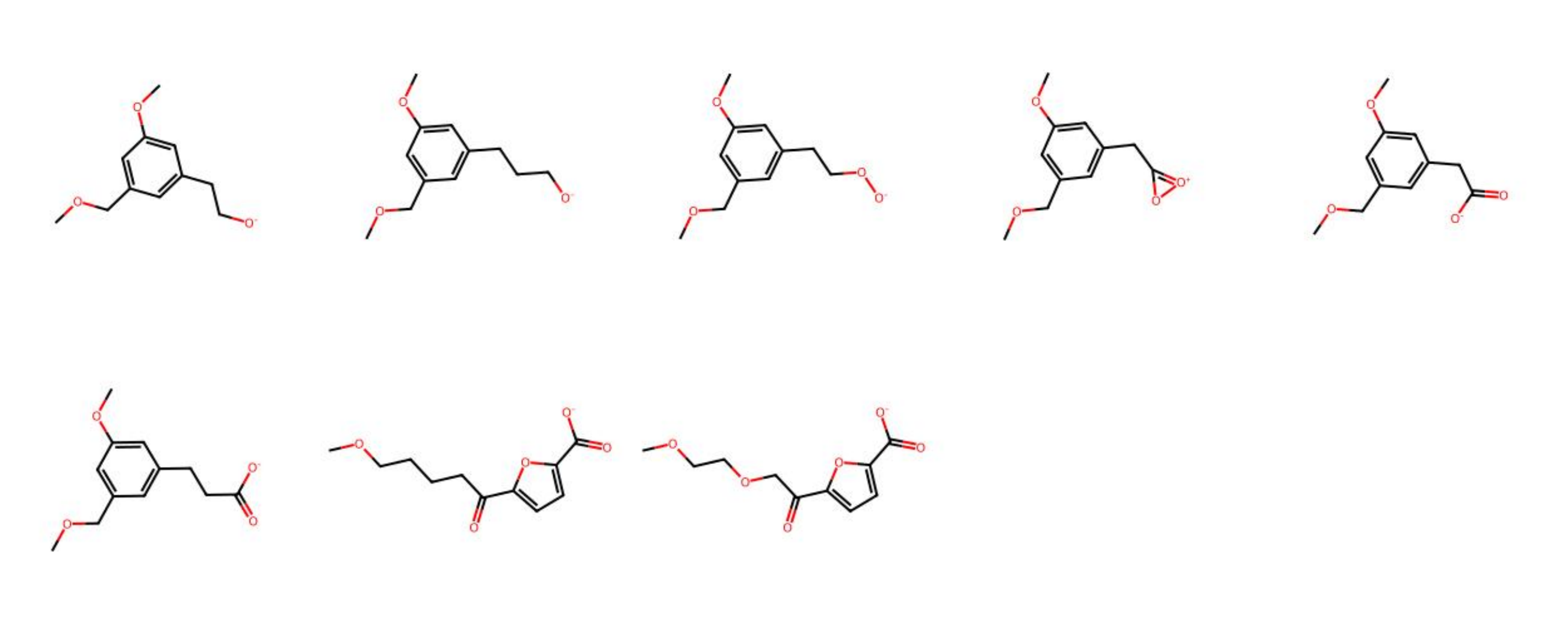}}
    \end{minipage}
    \hspace{0.01\textwidth}
    \begin{minipage}[c]{0.48\linewidth}
        \fbox{\includegraphics[width=0.9\linewidth]{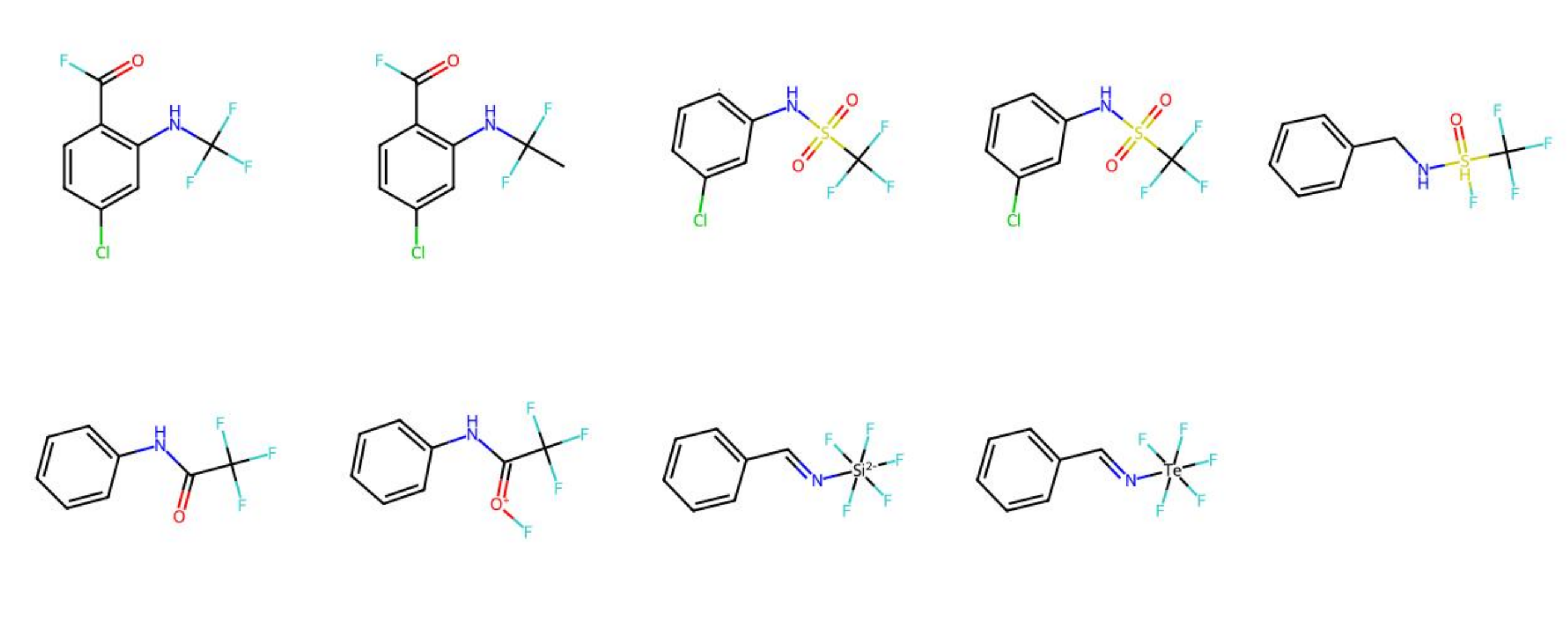}}
    \end{minipage}

    \caption{Interpolation between two pairs of molecules (only steps where the molecule is updated are shown). Starting molecules are top-left, progression is left to right, top to bottom.}
    \label{fig:interpolation}
\end{figure}

We also demonstrate that the latent embedding enables smooth transition between two molecules. To illustrate this, we perform linear and rotative interpolation directly in latent space between the embeddings of two valid molecules. This produces a continuous trajectory of latent vectors, each of which we decode back into a molecular graph. Because the decoder may map nearby latent points to identical structures, we retain only the distinct molecules encountered along the interpolation. Example interpolations sequences are shown in Figure \ref{fig:interpolation}, with more in Appendix \ref{app:interp}.

\paragraph{Local Property Optimization}

\begin{figure}[t]
    \centering
    \begin{minipage}{0.48\linewidth}
        \centering
    \fbox{\includegraphics[width=0.9\linewidth]{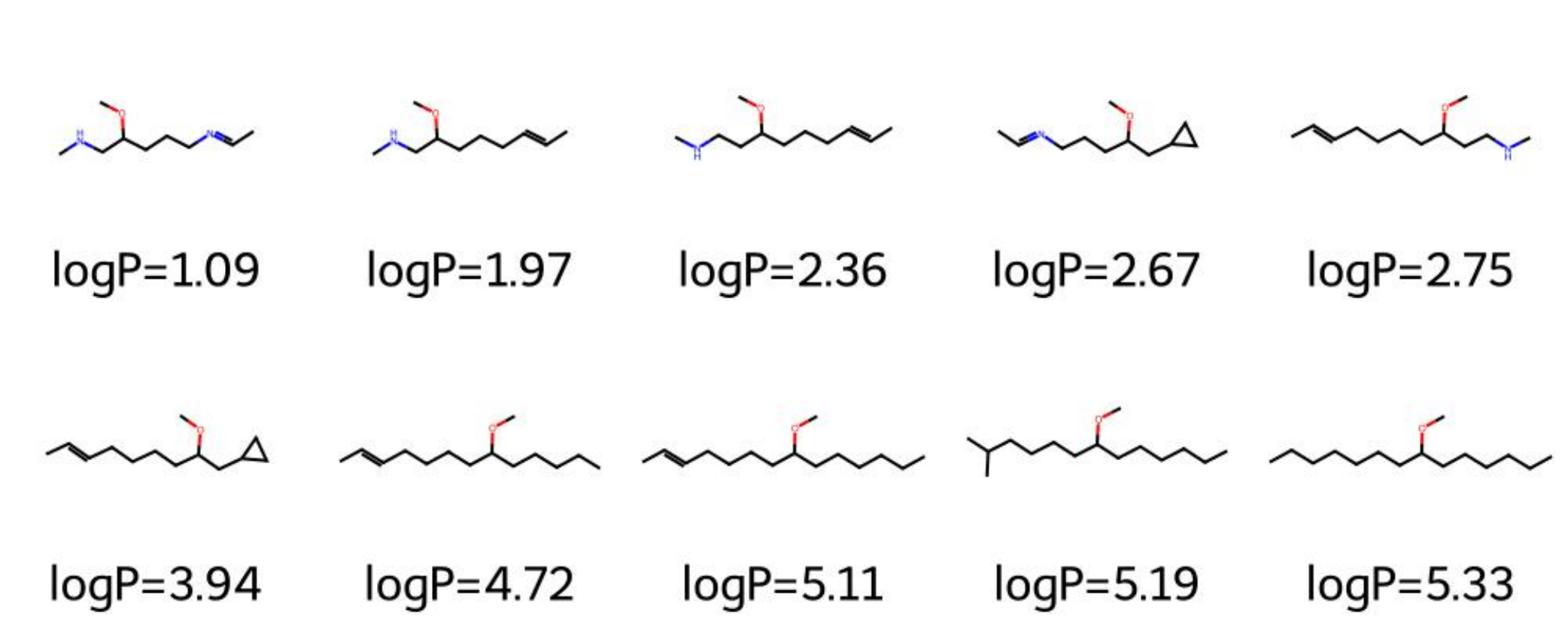}}
    \end{minipage}\hspace*{0.015\linewidth}
    \begin{minipage}{0.48\linewidth}
        \centering
        \fbox{\includegraphics[width=0.9\linewidth]{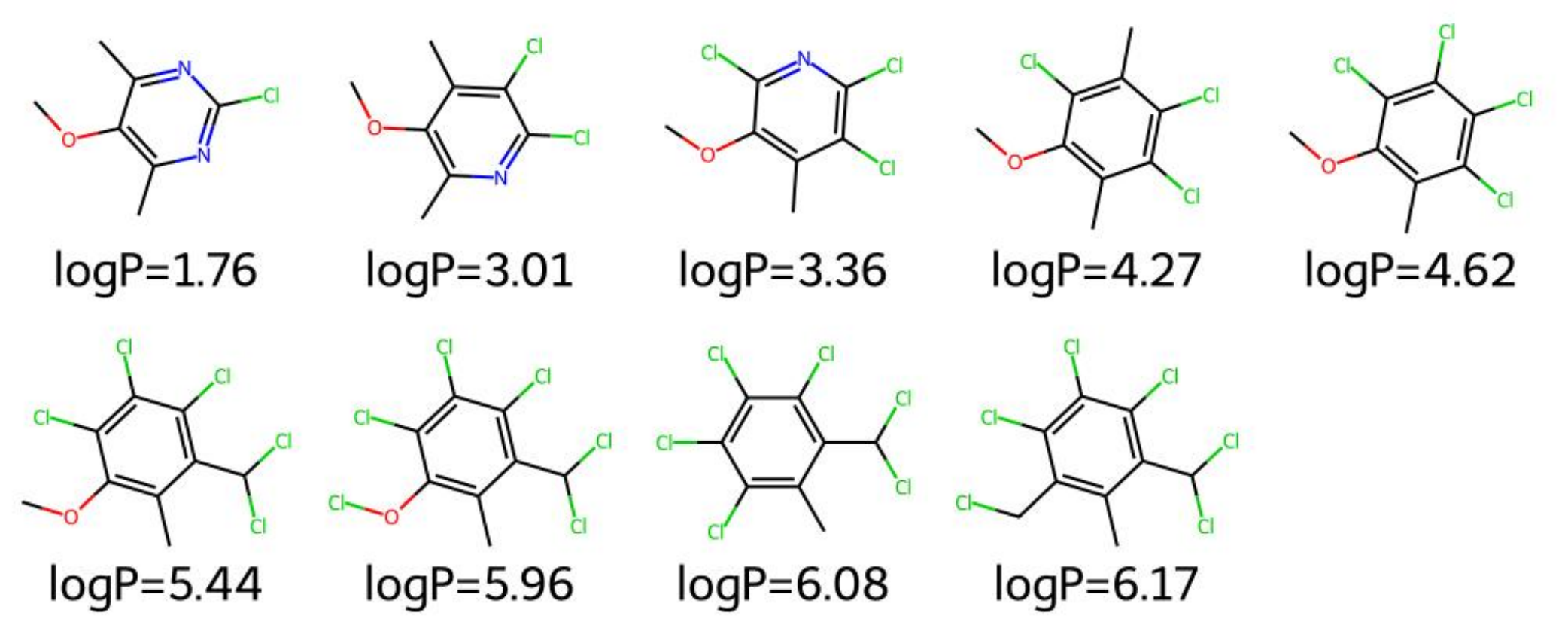}}
    \end{minipage}
    \caption{LogP maximization from 2 molecules. Starting from the embedding of the top-left molecule, the latent representation is optimized through CMA-ES to increase its logP (indicated below each molecule). Steps are displayed (left-to-right, top to bottom).}
    \label{fig:logpmax}
\end{figure}

We demonstrate that our latent embedding can be used not only to represent molecules but also to edit them in a controlled manner to improve a target property. To illustrate this capability, we perform black‑box optimization directly in the latent space using CMA‑ES \citep{hansen2019pycma,cmaes} to maximize logP \citep{logp1,logp2}, which quantifies the ratio of lipo- over hydro-philicity and is widely sought in drug design. Figure \ref{fig:logpmax} shows representative optimization trajectories.

The results highlight that the latent space supports smooth and minimal structural edits that incrementally improve the objective. CMA‑ES parameters enforce small steps between successive candidates, and we observe that the decoded molecules evolve gradually. We also notice the recovery of well‑established chemical patterns during optimization. In the left example, the algorithm increases logP by progressively lengthening the carbon chain. This behavior aligns with a documented principle: longer alkyl chains are increasingly hydrophobic, raising logP \citep{Mozrzymas2020,Yang2024,Xie_2021,SmithTanford1973Hydrophobicity}. On the right, the molecule develops a tetrachlorobenzene moiety, another structure associated with increased hydrophobicity \citep{MCPHEDRAN20132222}. The fact that such trends emerge purely from latent‑space navigation suggests that the model captures chemically meaningful directions that can be exploited for property‑guided molecular editing.

We also note that VAE performs significantly better than AE on this task, with the latter often not finding a near improvement beyond the original embedding, showing the benefits of imposing structure on the latent. On a batch of $1024$ molecules, average LogP improvement was $0.16$ using AE, and $2.72$ using VAE. Figure \ref{fig:ae_vae_optim} in Appendix \ref{app:logPmax} illustrates the performance on individual samples alongside more examples.

\paragraph{Graph generation capability}

\begin{table}[t]
    \caption{Generation performance of GraViti on QM9 and PubChem16, both in aromatic (A) and kekulized (K) form. Baselines for QM9 are reported directly from \citep{roos2026categoricalflowmaps}. VAE means direct latent decoding. VUN is the ratio of valid, unique and novel molecules (product of the three metrics).}
    \centering
    \resizebox{\textwidth}{!}{\begin{tabular}{l c c c c c c }
    \toprule
    \textsc{Dataset/Model} & \textsc{Steps} & \textsc{Valid. (\%)} & \textsc{Uniq. (\%)} & \textsc{Novelty (\%)} & \textsc{VUN (\%)} & \textsc{FCD}\\
    \midrule
    QM9(K)/CFM ECLD & 1 & 95.8 & 89.2 & N/A & 85.5 & 2.16\\
    QM9(K)/CFM CSD & 1 & 90.0 & 91.8 & N/A &82.6  & 2.14\\
    \midrule
    QM9(K)/GraViti & VAE & 93.2 & 98.5 & N/A & 91.8 & 2.88\\
    QM9(A)/GraViti & VAE & 92.0 & 99.0 & N/A & 91.1 & 3.00\\
    \midrule
    PubChem16(K)/GraViti & VAE & 68.1  & 100. & 91.7 & 62.4 & 4.20 \\
    PubChem16(A)/GraViti & VAE & 46.5  & 100. & 98.1 & 45.6 & 3.57 \\
    \bottomrule
    \end{tabular}}
    \label{tab:gen_perf}
\end{table}


{ While molecular graph generation is not the main focus of this work, it arises naturally from the architecture of a VAE. In particular, molecules can be generated in a single decoder pass by sampling the latent prior. We report generation results mainly as a diagnostic of the learned latent space and decoder, rather than as an attempt to optimize a dedicated molecular generator or establish a new state of the art, especially as VAEs are generally not competitive compared to more sophisticated, multi-step approaches such as diffusion \citep{conficlrVignacKSWCF23} or flow matching \citep{cheng2024categorical}.}

Table~\ref{tab:gen_perf} reports performance on QM9 and PubChem16 in
terms of validity, uniqueness, novelty, and Frechet ChemNet Distance \citep{10.1021/acs.jcim.8b00234}. Despite using no
chemistry-specific constraints during decoding, the non-aromatic QM9 model
achieves $93.18\%$ validity and $98.48\%$ uniqueness. These results are
competitive with recent methods specifically designed for one-step
generation. For example, one-step Categorical Flow Maps \citep{roos2026categoricalflowmaps} reports
$95.8\%$ validity and $89.2\%$ uniqueness with its ECLD objective, and
$90.0\%$ validity and $91.8\%$ uniqueness with its CSD objective. Our model therefore reaches a comparable validity regime while producing a larger fraction of unique samples.

Using explicit aromatic representation, we obtain 46.5\% validity on PubChem16 and 92\% on QM9. A large fraction of invalid molecules contain aromatic edges not forming closed aromatic cycles, reflecting the known difficulty of graph neural networks, including transformers, in representing cyclic structures without higher-order mechanisms \citep{WLTrans,WLTrans2,WLTrans3}. Training the model on the corresponding kekulized representation improves validity to 68\% on PubChem and 93\% on QM9. Examples of generated molecules are shown in Appendix \ref{app:gen_mols}.

We interpret these results as indicating that the latent space is locally well structured, enabling the generation of diverse and novel molecules in nearby regions, but globally insufficiently compact, leading to a significant fraction of invalid decodings when sampling more broadly. This observation is consistent with the literature on molecular VAEs \citep{Grammar,juncvae,NGUYEN20253867,Hu2023-xz}, which has shown that generation validity is commonly low in the absence of additional domain‑specific constraints \citep{Bhadwal2025,Ochiai2023}. Moreover, validity is known to be anti‑correlated with uniqueness and novelty in molecular generative models \citep{Mahmood2021}. In order to further probe the latent generative capabilities, we train a simple, cheap flow model in our latent space, raising validity to 89.7\% on PubChem16 in one step only (Appendix \ref{app:latentfm}).

\paragraph{Denoising} One application of autoencoding is removing noise from a perturbed input. In Appendix \ref{app:denoise}, we perturb molecules (up to 10 node and 10 edge labels are randomly changed), and show that we still recover valid molecules as output.

\paragraph{Regression} We further show that chemical and graph-level properties can be predicted directly from the latent space, including molecular quantum properties on QM9 and BIC scores on Bayesian networks (Appendix~\ref{app:reg}).

\section{Conclusion and future work}

We introduced \xxxx, a transformer‑based graph‑level variational autoencoder that learns compact latent representations of entire graphs and achieves state‑of‑the‑art reconstruction accuracy among graph-level VAEs on large graph datasets.

We demonstrate on molecular graphs that the latent space is coherent and chemically meaningful: it supports smooth interpolation, denoising, and property‑guided optimization, even recovering known chemical principles. Beyond molecular tasks, we also obtain strong performance on downstream inference problems for Bayesian graphs.

Finally, we show that a permutation invariant objective is not always required; when a stable node ordering is available, relaxing this constraint yields substantial gains in reconstruction quality and scalability. Future work includes integrating domain‑aware constraints to improve generation validity, exploring richer latent priors, and evaluating transfer to other graph domains such as circuits, proteins, and knowledge graphs. Overall, \xxxx\ offers a practical foundation for graph‑level representation learning and provides a usable latent space for downstream tasks.

\newpage
\bibliographystyle{unsrt}
\bibliography{sample}

\newpage
\appendix
\section{Design choices} \label{sec:designchoices}

\subsection{Graph-level or Node-level}

To motivate our use of a graph-level encoder (despite leveraging a canonical ordering), we implemented a node‑level version of our VAE, where the latent space consists of per‑atom embeddings. The architecture is the same as the main paper, but the readout and readin are removed. While this variant achieves higher reconstruction performance (99.94\% accuracy, 0.004 edit distance), its generative performance collapses: only 17\% of sampled molecules are valid, compared to 68\% for our graph‑level VAE.

This outcome is consistent with prior work showing that node‑wise generative models struggle to enforce global chemical constraints. GraphVAE reports at best 57\% validity on QM9 \citep{Simonovsky2018} (with the exception of a NoGM reaching 81\%, but due to a collapse on a mode, leading to only 24\% uniqueness). MolGAN \citep{gen2} also report that no baseline exceeds 57\% on QM9 (with the exception of NoGM), and uses an extensive reinforcement learning procedure specifically to improve validity. More recent analyses demonstrate that node‑level likelihoods fundamentally fail to capture global structural dependencies, with graph‑level modeling significantly improving validity: ``It is known from previous work that node-level representations, while useful for many applications, do not capture enough graph structure to generate realistic graphs" \citep{gen3}. In particular, the basic setting of sampling each node independently prevents sampling a globally coherent molecule with high probability.

Node‑level embeddings also break down outside the generation task. Because their dimensionality depends on the number of atoms, they live in different latent spaces for different molecules, so there is no canonical way to define distances or interpolate between them. This makes latent‑space editing, smooth interpolation, and property‑guided traversal ill‑posed: one cannot move ``between" molecules when their latent representations have incompatible shapes. In the same way, optimization algorithms such as CMA-ES work on fixed-sized spaces. Graph‑level embeddings avoid this entirely by mapping every molecule into a single fixed‑dimensional latent space where distances, directions, and edits are well defined, regardless of size.

\begin{table}[H] \centering \caption{Comparison between graph-level and node-level latent representations.} \label{tab:graph_vs_node} \begin{tabular}{lccc} \toprule Model & Reconstruction Acc. ($\uparrow$) & Edit Dist. ($\downarrow$) & Validity (\%, $\uparrow$) \\ \midrule Graph-level VAE & 99.50 & 0.04 & 67 \\ Node-level VAE & 99.94 & 0.004 & 17 \\ \bottomrule \end{tabular} \end{table}

\subsection{Permutation-invariant encoder} \label{annex:pie}

Since our loss function is permutation-sensitive, it could be assumed that our model would benefit from a permutation-sensitive encoder. We justify here the choice for a permutation-invariant encoder.

In order to test that theory, we experiment on PubChem16 replacing the RW- and Laplacian- positional encodings (both permutation equivariant) with a sinusoidal positional encoding (similar to input of the decoder, permutation-sensitive). The encoder, and therefore latent representation are thus a sequence-model, where different node orderings can result in different representations. Our results are presented in table \ref{tab:PEvsPI}.

\begin{table}[H]  \centering \caption{Comparison between a permutation-invariant and a permutation-sensitive encoder (PubChem16).} \label{tab:PEvsPI} 
\begin{tabular}{lcc} 
\toprule Model & Reconstruction Acc. ($\uparrow$) & Edit Dist. ($\downarrow$) \\ 
\midrule 
Default & 99.50 & 0.04 \\ 
Permutation-Sensitive & 99.95 & 0.001 \\ 
\bottomrule 
\end{tabular} 
\end{table}

We observe that injecting the node-ordering information makes reconstruction easier for our model. We tested the robustness of our permutation-sensitive variant by evaluating its reconstruction capacity on molecules obtained by random node permutations of the input. None of the reconstructed graphs matched their input counterparts, showing that the learned representation is not only not permutation-equivariant, but more generally incapable of properly encoding any graph that is not provided in the canonical ordering. In particular, $65\%$ of the reconstructed graphs gave invalid molecules (out of a sample of 2048 molecules).

 However, this gain is conditional on the ordering convention being reliably available. In the case where the ordering exists as an external, callable function, this is not an issue, since input can be pre-treated to be in the right ordering. On the other hand, if the ordering is a property of the training dataset but not otherwise accessible (as is the case with QM9 and PubChem, which both use closed-source canonization) the permutation-invariant model can still train and generalize (since the canonical ordering is only required for loss computation and can be ignored at inference), but the permutation-sensitive variant cannot function properly.

\newpage
\section{Statistics of the datasets} \label{app:datasets}

\subsection{Statistics}

\begin{table}[h]
    \centering
    \caption{Statistics of the used datasets.}
    \begin{tabular}{ c c c c c c }
        \toprule
         Name & \# Graph & Avg $\#$ nodes & Avg $\#$ edges & Max Size & Node types \\
         \midrule
         QM9NoHydro & $130$k & $8.79$ & $9.4$ & $9$ & $4$\\
         PubChem16 & $12.89$M & $13.87$ & $14.19$ & $16$ & $31$\\
         PubChem32 & $74.34$M & $21.75$ & $21.15$ & $32$ & $31$\\
         Coloring & $320$k & $12.51$ & $24.93$ & $20$ & $4$ \\
         Asia (BN) & $200$k & $8$ & $8.05$ & $8$ & $8$ \\
         \bottomrule
    \end{tabular}

    \label{tab:datasets}
\end{table}

\subsection{Canonizations}

All datasets below are publicly provided in canonized form, and were used as-is.

QM9 \citep{qm9} is a subset of a larger database, GDB‑17. GDB‑17\footnote{\url{https://gdb.unibe.ch/downloads/}} \citep{gdb} is stored as canonized SMILES strings.

The canonization of GDB‑17's SMILES representations is documented in the primary enumeration paper \citep{gdb}. The methods section states that all software was implemented using JChem libraries from the ChemAxon proprietary software, and that each generated molecule was converted to a unique SMILES string specifically to detect and remove duplicates before storage. Note that the paper uses the term ``unique" rather than ``canonical" when describing this step.

PubChem's canonization applies directly to atom ordering rather than only to an upstream SMILES string. PubChem's standardization pipeline invokes OpenEye's OECanonicalOrderAtoms as part of its aromaticity and kekulized form standardization\footnote{\url{https://docs.eyesopen.com/toolkits/python/oechemtk/OEChemFunctions/OECanonicalOrderAtoms.html}} \citep{Hahnke2018PubChem}. The resulting canonical SMILES strings are parsed with RDKit's Chem.MolFromSmiles function, which preserves the input order rather than recomputing its own canonization, so the canonical ordering established by OpenEye is retained in the molecules we construct.

Neither OpenEye's OEChem nor ChemAxon's JChem are open source, so we do not have access to the specific canonization algorithms in either case. This is not an obstacle in our case thanks to the encoder being permutation invariant. As long as the training dataset's ordering remains consistent (as it is), the model functions well on any new molecule, regardless of ordering, so long as no permutation-sensitive loss is required.

\newpage
\section{Hyperparameters and training details}\label{app:training}

We provide in table \ref{tab:hyperparam} the hyperparameters used to train our models. Our models use a half-sinusoidal warmup curve for beta (KL-divergence) and a cosine annealing scheduler with warmup and restarts.

\begin{table}[H]
    \caption{Hyperparameters used for best models. All experiments were conducted on AMD MI210 GPUs.}\label{tab:hyperparam}
    \centering
    \begin{tabular}{l c c c c c}
        \toprule
        Dataset & QM9 & PubChem16 & PubChem32 & Coloring & Asia (BN) \\
        \midrule
        Optimizer & AdamW & AdamW & AdamW & AdamW & AdamW \\
        Learning rate & 2e-4 & 1e-4 & 1e-4 & 1e-4 & 1e-4 \\
        Batch size & 64 & 64 & 256 & 64 & 64 \\
        Latent dimension & 64 & 256 & 512 & 128 & 64 \\
        Encoder hidden size & 128 & 512 & 512 & 256 & 64 \\
        Encoder input size  & 128 & 128 & 128 & 128 & 64 \\
        Graph transformer layers & 2 & 5 & 7 & 3 & 1 \\
        Transformer heads & 4 & 4 & 8 & 4 & 1 \\
        Beta (VAE) & 1e-5 & 1e-4 & 1e-5 & 1e-4 & 1e-3 \\
        LR cycle length & 200 & 200 & 800 & 300 & 30 \\
        Beta cycle length & 400 & 200 & 800 & 200 & 30\\
        Ratio negative edges & 4 & 4 & 4 & 4 & 4 \\
        Edge loss weight & 1.25 & 1.25 & 1.25 & 1 & 1 \\
        Train split & 0.8 & 0.9736 & 0.9978 & 0.9375 & 0.9 \\
        Valid. split & 0.1 & 0.003 & 0.0011 & 0.03125 & 0.05 \\
        Test. split & 0.1 & 0.0234 & 0.0011 & 0.03125 & 0.05 \\
        GPUs (training) & 1 & 1 & 2 & 1 & 1 \\
        GPUs (inference) & 1 & 1 & 1 & 1 & 1 \\
        \bottomrule
    \end{tabular}
\end{table}

\newpage
\section{Performance on molecules in aromatic form} \label{app:arom}

\begin{table}[h]
\caption{Percentage of perfectly reconstructed graphs and edit distance. Results from PIGVAE are reported from \citep{grale}.}
\label{tab:recon_perf_arom}
\centering
{\begin{tabular}{l c c c c c}
\toprule
\textsc{Dataset} & \textsc{Metric} & \textsc{GRALE} & \textsc{PIGVAE}$^*$ & \textsc{GraViti (AE)} & \textsc{GraViti (VAE)}
 \\
\midrule
\multirow{2}{*}{QM9NoHydro}
  & Edit Dist. & \textbf{0.05} & N/A & 0.10 & 0.10\\
  & Accuracy   & {97.2} & N/A & 97.5 & \textbf{97.7}\\
\midrule
\multirow{2}{*}{PubChem16}
  & Edit Dist. & 0.06 & 1.69$^*$ & \textbf{0.05} & \textbf{0.06} \\
  & Accuracy   & 96.8 & 41.0$^*$ & \textbf{99.3} & \textbf{99.3} \\
\midrule
\multirow{2}{*}{PubChem32}
  & Edit Dist. &  0.70 & 2.53$^*$ & 0.17 & 0.18 \\
  & Accuracy   &  73.3 & 24.9$^*$ & 98.0 &  97.3 \\
\bottomrule 
\end{tabular}}
\end{table}

\newpage
\section{Loss function} \label{sec:loss}

\subsection{Reconstruction Loss}

{The reconstruction loss is defined over nodes, edges and predicted size. Since both nodes and edges take discrete values (atom and bond types), we treat our graph prediction task as multiple classification tasks, one per node and edge. We use per-node and per-edge cross-entropy:

\[\mathcal{L}_\texttt{CE}(p,y) = -\sum_{i=1}^{C} y_{i} \log(p_i)\]

with $p_i$ the predicted probability of class $i$ and $y$ a one-hot vector for true class. Despite facing a highly imbalanced distribution of classes, with some elements appearing orders of magnitude more often than others, we found that cross-entropy with inverse-frequency weights and focal loss \citep{focalloss} consistently underperformed naive cross-entropy. For a graph with $N$ nodes, with $y^n$ (resp. $y^{n_1, n_2}$) the one-hot encoding of classes for node $n$ (resp. edge $n_1, n_2$) and $p^n$ (resp. $p^{n_1, n_2}$) the predicted class probabilities for this node (resp. edge), we therefore have}:

\[\mathcal{L}_\texttt{node} = \frac{1}{N}\sum_{i=1}^N \mathcal{L}_\texttt{CE}(p^{i}, y^i) \text{ and } \mathcal{L}_\texttt{edge} = \frac{1}{N^2}\sum_{i=1}^N\sum_{j=1}^N \mathcal{L}_\texttt{CE}(p^{i,j}, y^{{i,j}})\]

In order to avoid the overwhelmingly frequent ``no-edge" class from hurting the training, we actually sample a number of ``no-edge" class equal to $rN_e$, with $r$ a hyperparameter and $N_e$ the number of actual edges.

A natural concern is that this loss is applied in the input node order, which makes it highly sensitive to permutations. As we show in the section \ref{sec:exp}, this inductive bias is beneficial for dataset with a consistent canonical ordering: enforcing a consistent ordering improves reconstruction fidelity and does not hinder the model’s ability to represent graphs up to isomorphism. Finally, the error $\mathcal{L}_{\text{size}}$ on the predicted log-size $\hat{s}$ is simply the smooth $L1$ error with the ground truth log-size $s$. Our final reconstruction loss is a combination of those three losses:

\[
\mathcal{L}_{\text{rec}}
= \mathcal{L}_{\text{node}}
+ \eta \mathcal{L}_{\text{edge}}
+ \mathcal{L}_{\text{size}} \text{ with $\eta$ a hyperparameter.}
\]

\subsection{Variational loss}

For the latent variables we assume a standard normal prior
\(\mathcal{N}(0, I)\).
Given encoder outputs \(z_\mu\) and \(z_\sigma\),
the variational term corresponds to the KL divergence
between the approximate posterior
\(q(z \mid x) = \mathcal{N}(z_\mu, \mathrm{diag}(z_\sigma^2))\)
and the prior:

\[
\mathcal{L}_{\text{KL}}
= D_{\mathrm{KL}}\!\left(
    \mathcal{N}(z_\mu, \mathrm{diag}(z_\sigma^2))
    \,\Vert\,
    \mathcal{N}(0, I)
\right)
=
-\frac{1}{2}
\sum_{i=1}^d
\left(
    1 + \log z_{\sigma,i}^2 - z_{\mu,i}^2 - z_{\sigma,i}^2
\right).
\]

The final training objective is therefore:

\[
\mathcal{L}
= \mathcal{L}_{\text{rec}}
+ \beta\,\mathcal{L}_{\text{KL}},
\]

where \(\beta\) optionally controls the strength of the variational term. In practice, $\beta$ follows a warmup schedule, sinusoidally increasing until plateauing at its maximum value.

\paragraph{Extra loss terms for molecules}
In the case of { molecules in aromatic form}, the graph representation (atom-types/bond-type) does not contain all of the relevant information when using the aromatic representation. To alleviate this ambiguity, we also predict the formal charge and number of bonded hydrogen atoms for all atoms that are part of an aromatic ring using two extra prediction heads.

\newpage
\section{Order enforced on COLORING} \label{sec:orderingcolor}

To demonstrate the importance of ordering for our model, we also run an ablation study on the COLORING dataset \citep{krzakala2024endtoend}. This dataset contains general graphs with only node classes, and such that two nodes of the same color cannot be connected. As such, there is no inherent ordering of the node. To enforce ordering on a Coloring graph $G = (V, E)$, we replace the arbitrary node order with a deterministic WL-based order. Let $c_v$ the sampled color label of node $v$. We initialize the WL labels as $h_v^{(0)} = c_v$. For $t = 1, 2, 3$, each node receives the signature
\[
    s_v^{(t)} = \left (h_v^{(t-1)}, sort\left ({h_u^{(t-1)} : u \in N\left (v\right )}\right )\right ),
\]
where $N(v)$ is the set of graph neighbors of $v$. The set of signatures is then compressed into integer labels $h_v^{(t)}$ by sorting the unique signatures lexicographically and assigning consecutive IDs. Thus, after three iterations, $h_v^{(3)}$ encodes the color and topology of the node's local neighborhood up to radius three, as distinguished by 1-WL refinement.

The final node order is obtained by sorting nodes lexicographically using the tuple
\[
    \left (h_v^{(3)}, d_r\left (v\right ), d_a\left (v\right ), deg\left (v\right ), y_v, x_v\right ),
\]
where $d_r(v), d_a(v)$ are the unweighted shortest-path distances from selected root and anchor nodes, $deg(v)$ is the graph degree, and $(x_v, y_v)$ is the geometric position of the node. 

The root $r$ is chosen as the smallest structurally identifiable node after WL refinement, i.e. the node minimizing
\[
    \left (h_v^{(3)}, deg\left (v\right ), y_v, x_v\right ).
\]

The anchor $a$ is chosen far from the root to further break symmetries by adding a second axis, as the node maximizing
\[
    \left (d_r\left (v\right ), deg\left (v\right ), -h_v^{(3)}\right ).
\]

\newpage
\section{Latent optimization of LogP} \label{app:logPmax}

We show below some extra examples for optimizing logP as described in section \ref{sec:exp}.

\begin{figure}[H]
    \centering
    \begin{subfigure}[t]{0.48\linewidth}
        \centering
        \includegraphics[width=\linewidth]{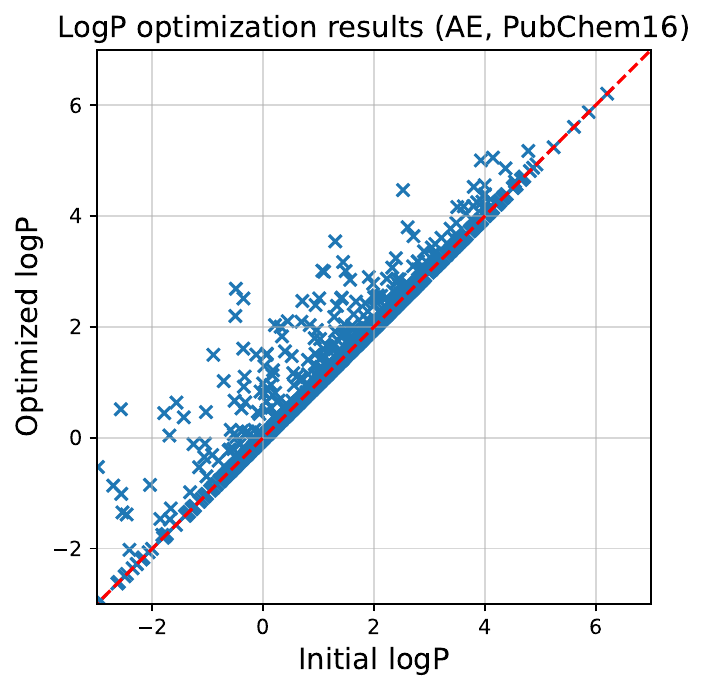}
        \caption{AE optimization}
        \label{fig:ae_optim}
    \end{subfigure}
    \hfill
    \begin{subfigure}[t]{0.48\linewidth}
        \centering
        \includegraphics[width=\linewidth]{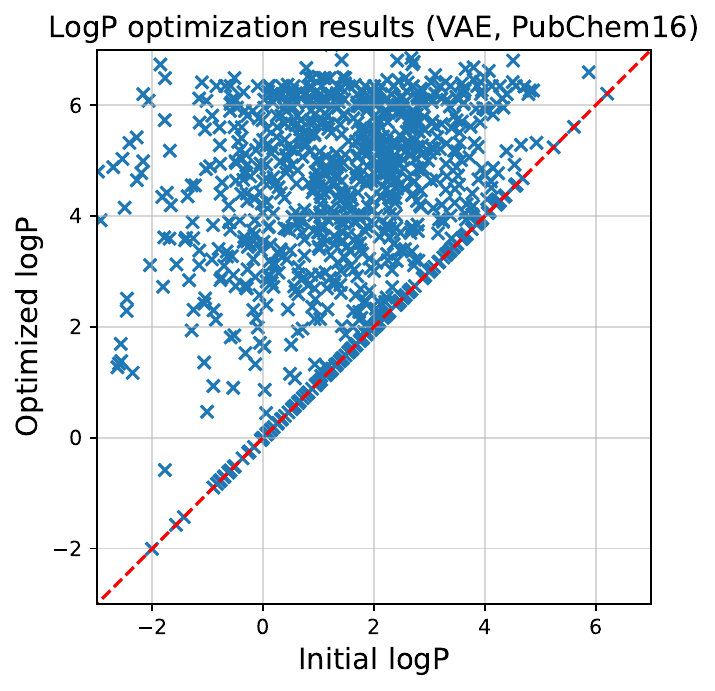}
        \caption{VAE optimization}
        \label{fig:vae_optim}
    \end{subfigure}
    \caption{Comparison of AE and VAE optimization performance on 128 random molecules. The structure in the VAE latent allows significantly better improvement.}
    \label{fig:ae_vae_optim}
\end{figure}

CMA-ES is obtained from the \texttt{cma} package \footnote{https://pypi.org/project/cma/} run for 100 steps, with \texttt{sigma0} set to 0.25, 50 iterations, and a population size of 16.
\newpage

\begin{figure}[H]
    \centering
    \fbox{\includegraphics[width=0.9\linewidth]{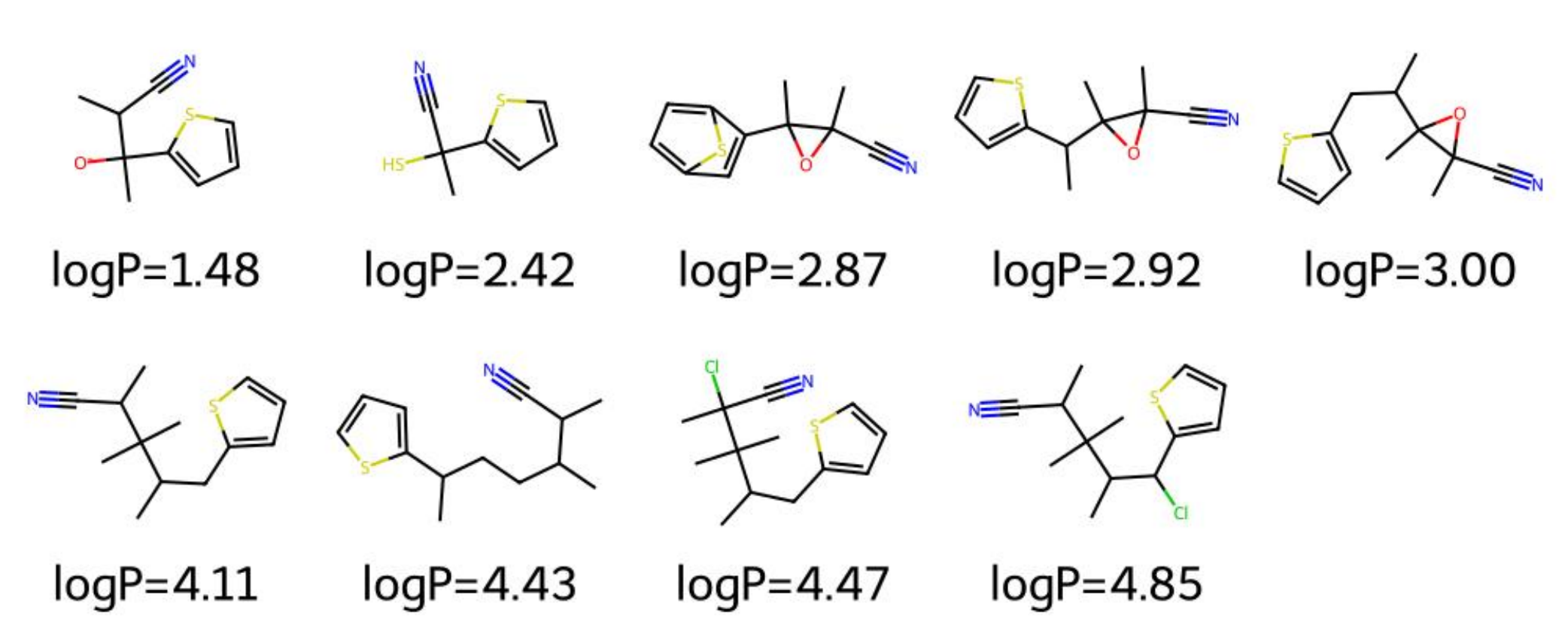}}
    \fbox{\includegraphics[width=0.9\linewidth]{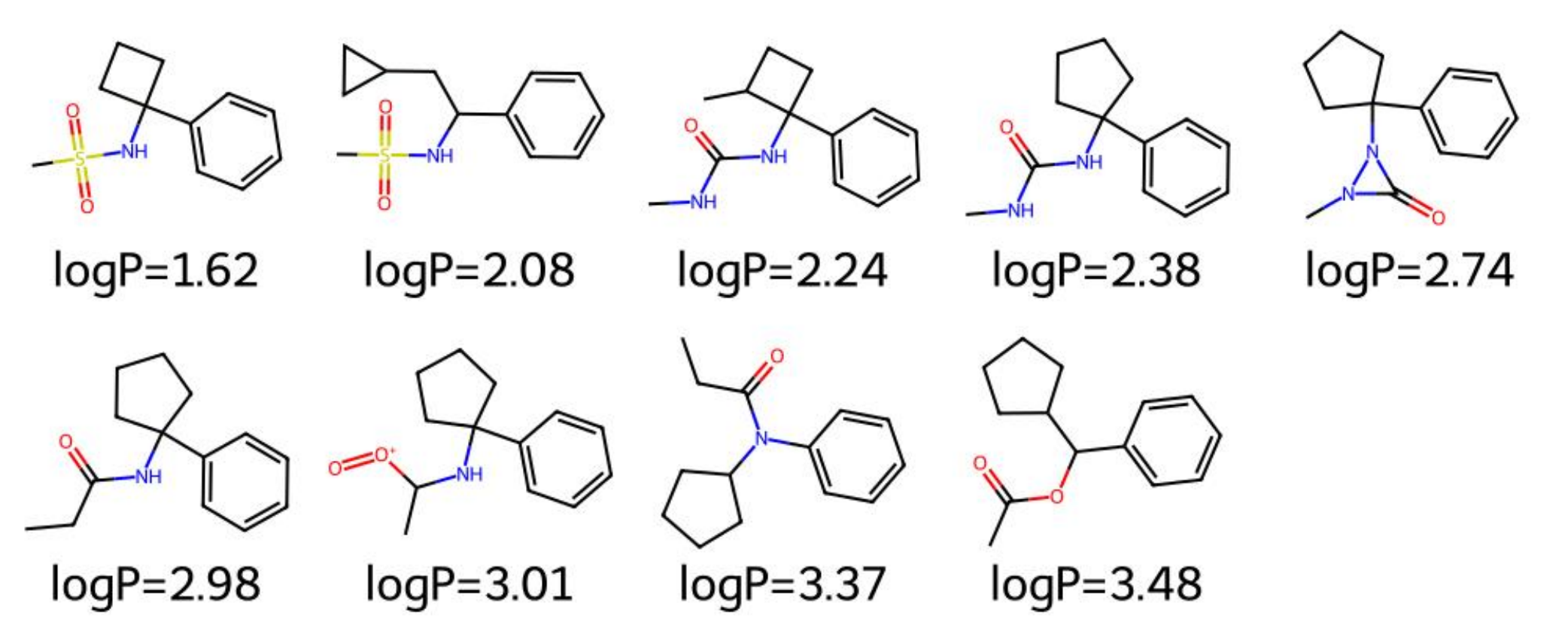}}
    \centering
    \fbox{\includegraphics[width=0.9\linewidth]{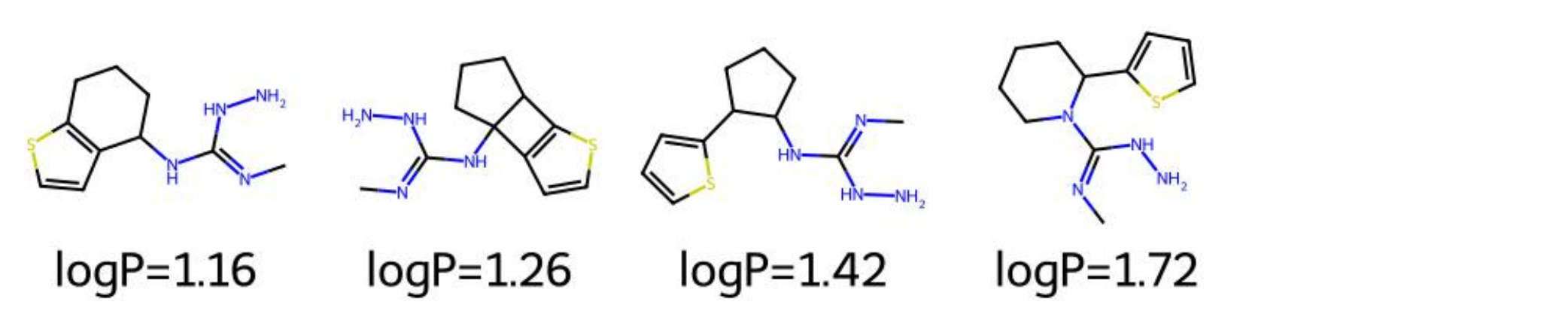}}
    \centering
    \fbox{\includegraphics[width=0.9\linewidth]{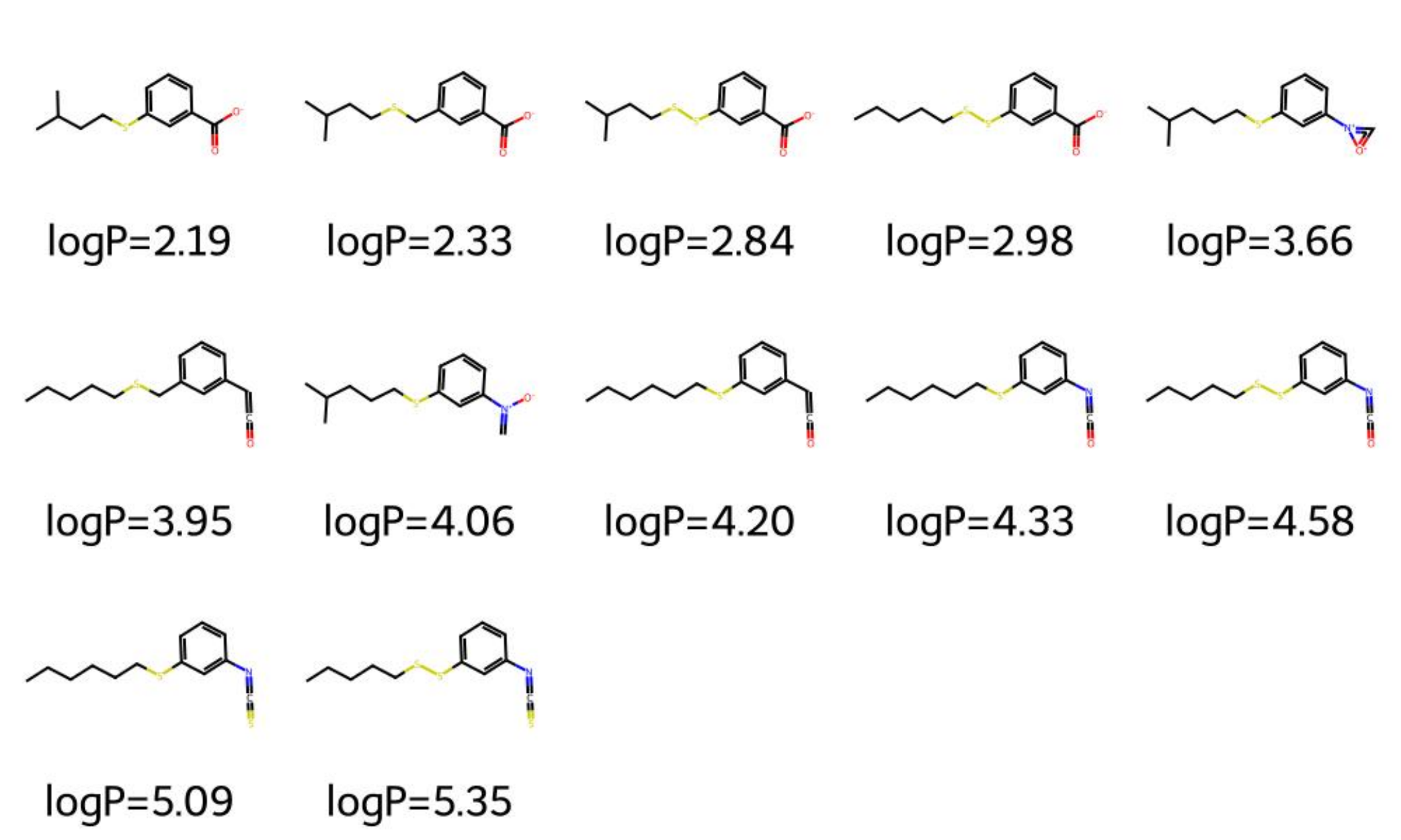}}
    \caption{Extra logP optimization plots (PubChem16).}
\end{figure}

\begin{figure}[H]
    \centering
    \fbox{\includegraphics[width=0.9\linewidth]{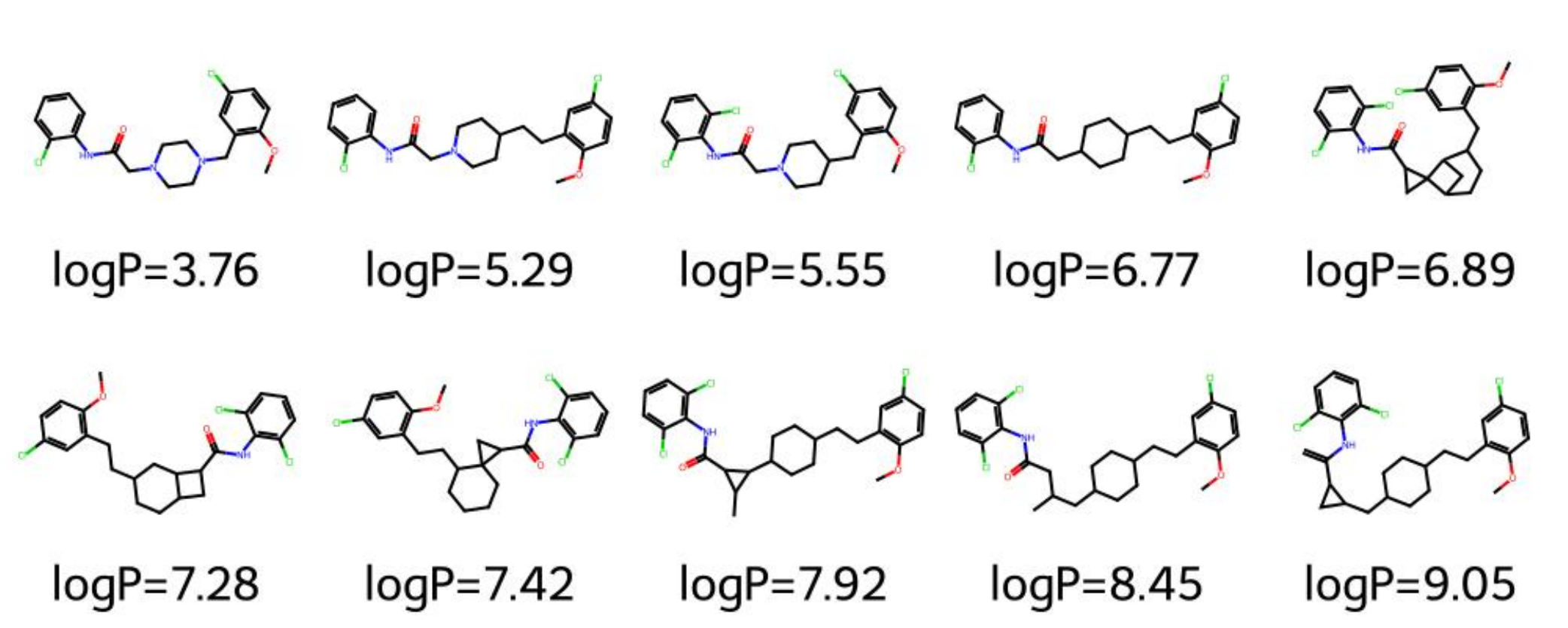}}
    \fbox{\includegraphics[width=0.9\linewidth]{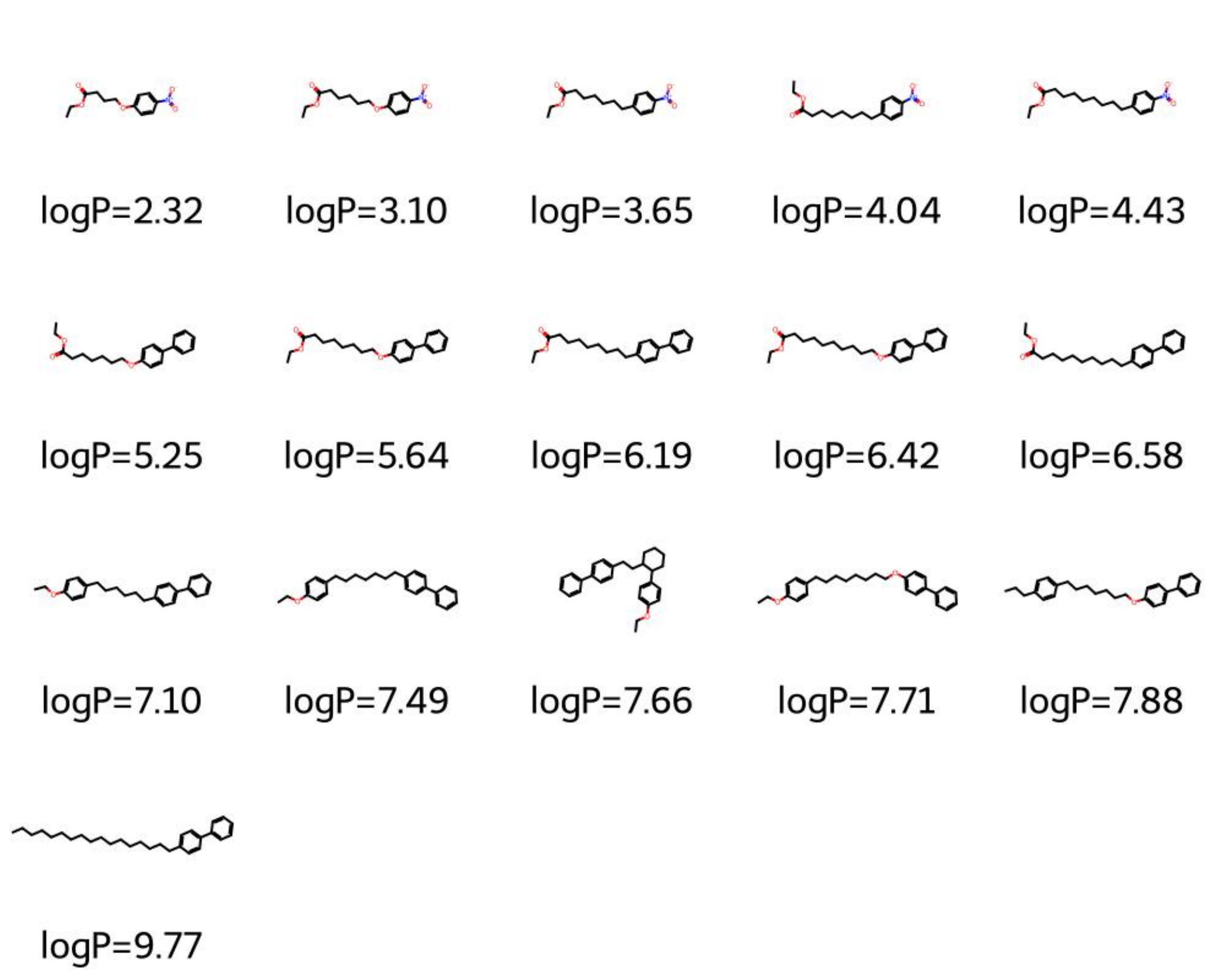}}
    \centering
    \fbox{\includegraphics[width=0.9\linewidth]{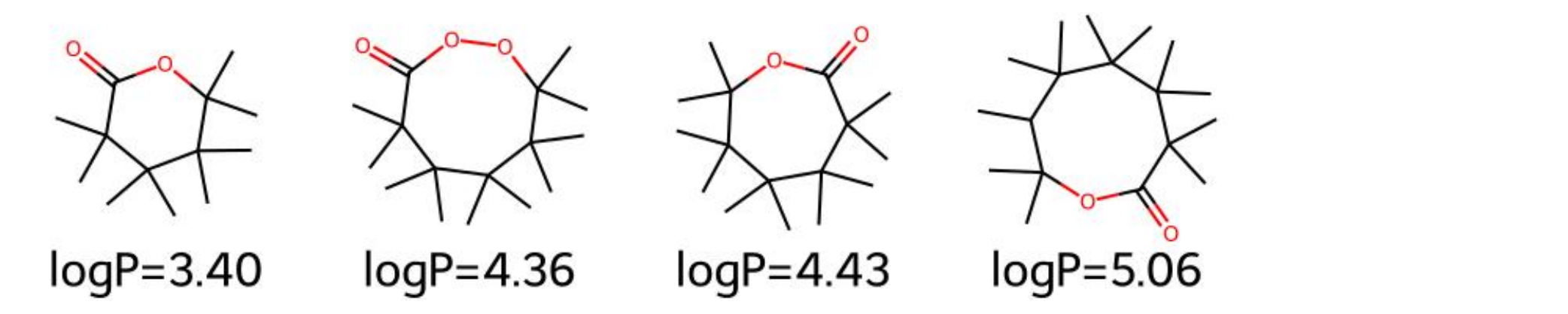}}
    \centering
    \fbox{\includegraphics[width=0.9\linewidth]{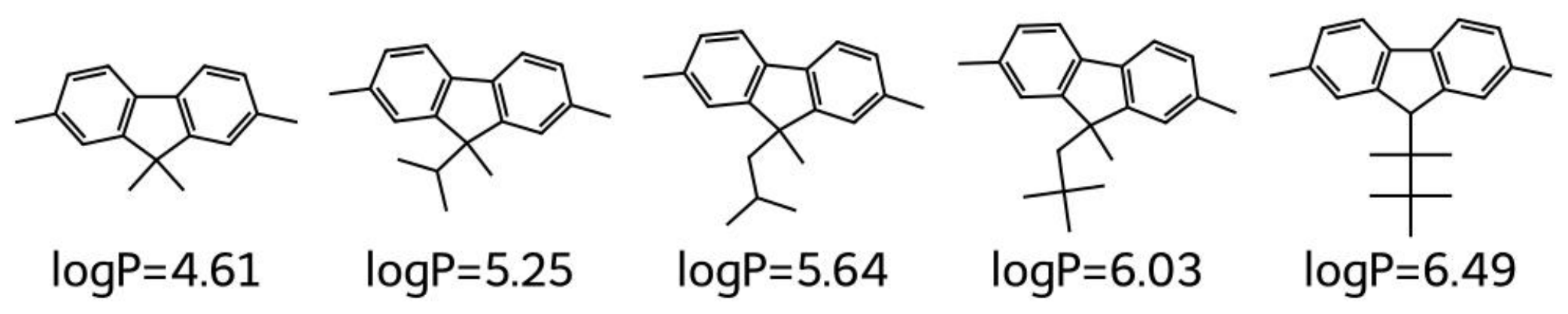}}
    \caption{Extra logP optimization plots (PubChem32).}
\end{figure}

\newpage

\section{Downstream Regression} \label{app:reg}

To evaluate the chemical expressiveness and structural organization of our latent space, we test whether meaningful molecular information can be recovered directly from the embeddings. We embed the QM9 molecules using a model trained on PubChem. Using only the latent vectors as input, we train simple regressors to predict three key molecular properties: dipole moment, HOMO–LUMO gap, and $U_0$ the 0 K internal energy. These properties probe complementary aspects of molecular structure and chemical behavior. The results are reported in table \ref{tab:downstream}.

\begin{table}[H]
    \centering
    \caption{Downstream task: regression on molecular properties. MSE is reported (mean / standard dev. over 5 folds). Models are trained either on PubChem16 or 32, as indicated.}
    \begin{tabular}{l c c c }
    \toprule
    \textsc{Model} & \textsc{Dipole moment} & \textsc{Homo-Lumo} & \textsc{Energy at 0 Kelvin} \\
    \midrule
    GRALE (PC16) & $0.4029 \pm 0.0071$ & $0.1230 \pm 0.0026$ & $\textbf{0.0033 $\pm$ 0.0010}$ \\
    GRALE (PC32) & $0.3677 \pm 0.0105$ & $0.1484 \pm 0.0033$ & $0.0204 \pm 0.0024$ \\
    \midrule
    GraViti (VAE, PC16) & $0.3219 \pm 0.0042$ & $0.0975 \pm 0.0032$ & $0.0056 \pm 0.0001$ \\
    GraViti (AE, PC16) & $0.3322 \pm 0.0016$ & $0.0929 \pm 0.0013$ & $0.0035 \pm 0.0002$ \\
    GraViti (VAE, PC32) & $\textbf{0.2937 $\pm$ 0.0012}$ & $0.0861 \pm 0.0030$ & $0.0106 \pm 0.0053$ \\
    GraViti (AE, PC32) & ${0.3040} \pm 0.0096$ & $\textbf{{0.0853} $\pm$ 0.0017}$ & $0.0034 \pm 0.0002$ \\
    \bottomrule
    \end{tabular}
    \label{tab:downstream}
\end{table}

To further evaluate whether the latent space captures graph-level information beyond molecular structure, we consider a regression task on Bayesian networks. We use the BN dataset introduced in D-VAE \citep{zhang2019dvaevariationalautoencoderdirected}, consisting of 200{,}000 8-node Bayesian networks generated with the bnlearn package \citep{JSSv035i03}. Each graph is assigned a Bayesian Information Criterion (BIC) score measuring how well the network structure fits the Asia dataset \citep{10.1111/j.2517-6161.1988.tb01721.x}. We first train small models for reconstruction (where all models and baseline reach $100\%$ accuracy), and using the encoder embeddings we train MLP regressors to predict the BIC score. The results are reported in table \ref{tab:bn-downstream}.

\begin{table}[H]
    \centering
    \caption{Downstream task: regression on BIC score. RMSE and Pearson correlation coefficient are reported (mean and standard deviation). Models are trained on BN dataset.}
    \begin{tabular}{l c c }
    \toprule
    \textsc{Model} & \textsc{RMSE} & \textsc{Pearson's $r$} \\
    \midrule
    GRALE & $0.0247 \pm 0.0052$ & $\textbf{0.9997 $\pm$ 0.0001}$ \\
    \midrule
    GraViti (VAE) & $0.0457 \pm 0.0021$ & $0.9990 \pm 0.0001$ \\
    GraViti (AE) & $\textbf{0.0229 $\pm$ 0.0038}$ & $\textbf{0.9997 $\pm$ 0.0001}$ \\
    \bottomrule
    \end{tabular}
    \label{tab:bn-downstream}
\end{table}

\newpage

\section{Denoising}\label{app:denoise}

\begin{figure}[H]
    \centering
    \includegraphics[width=\linewidth]{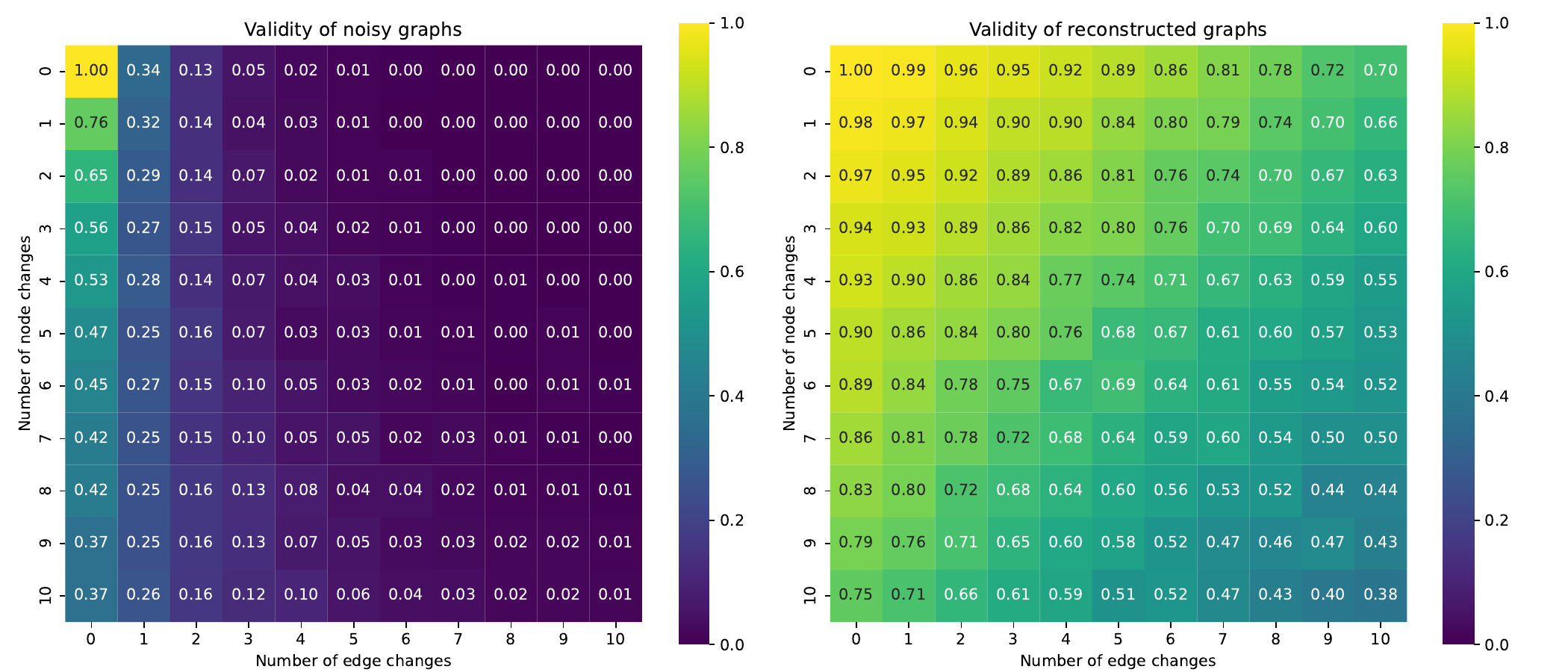}
    \caption{Denoising results: each cell reports the fraction of valid molecules (1024 molecules from PubChem16) after applying a number of random node and edge edits (left) and after reconstruction by the model (right). The AE consistently restores chemically valid structures even under heavy perturbations.}
    
    \includegraphics[width=\linewidth]{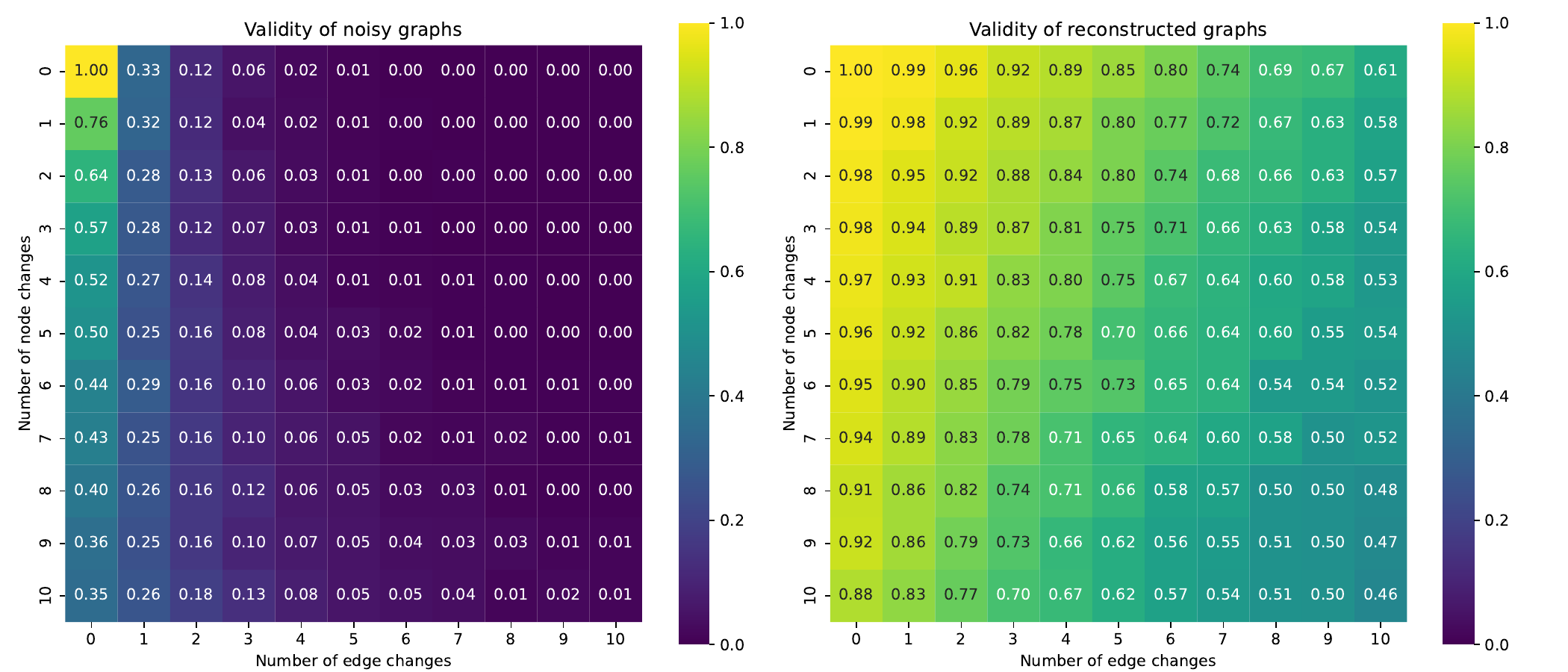}
    \caption{Denoising results: each cell reports the fraction of valid molecules (1024 molecules from PubChem16) after applying a number of random node and edge edits (left) and after reconstruction by the model (right). The VAE consistently restores chemically valid structures even under heavy perturbations.}
    \label{fig:denoiseaevae}
\end{figure}

Autoencoders are effective tools for denoising structured data. To evaluate this capability, we sample 1024 graphs from PubChem16 and introduce controlled noise by randomly altering node and edge labels, which can include the creation or deletion of edges. We then forward these corrupted graphs through our model. Figure \ref{fig:denoiseaevae} reports the validity of molecules after the noising procedure and after reconstruction by the autoencoder. The results show a clear improvement in molecular validity once the noisy inputs are passed through the model, confirming its denoising ability, even with large numbers of edits.

Fig. \ref{fig:denoiseaevae} compares the denoising abilities  of AE (top) and VAE (bottom), showing a slight advantage for the variational model.

\newpage
\section{Molecule interpolation} \label{app:interp}

We show below some extra examples for molecule interpolation described in Section \ref{sec:exp}.

\begin{figure}[H]
    \centering
    \fbox{\includegraphics[width=0.9\linewidth]{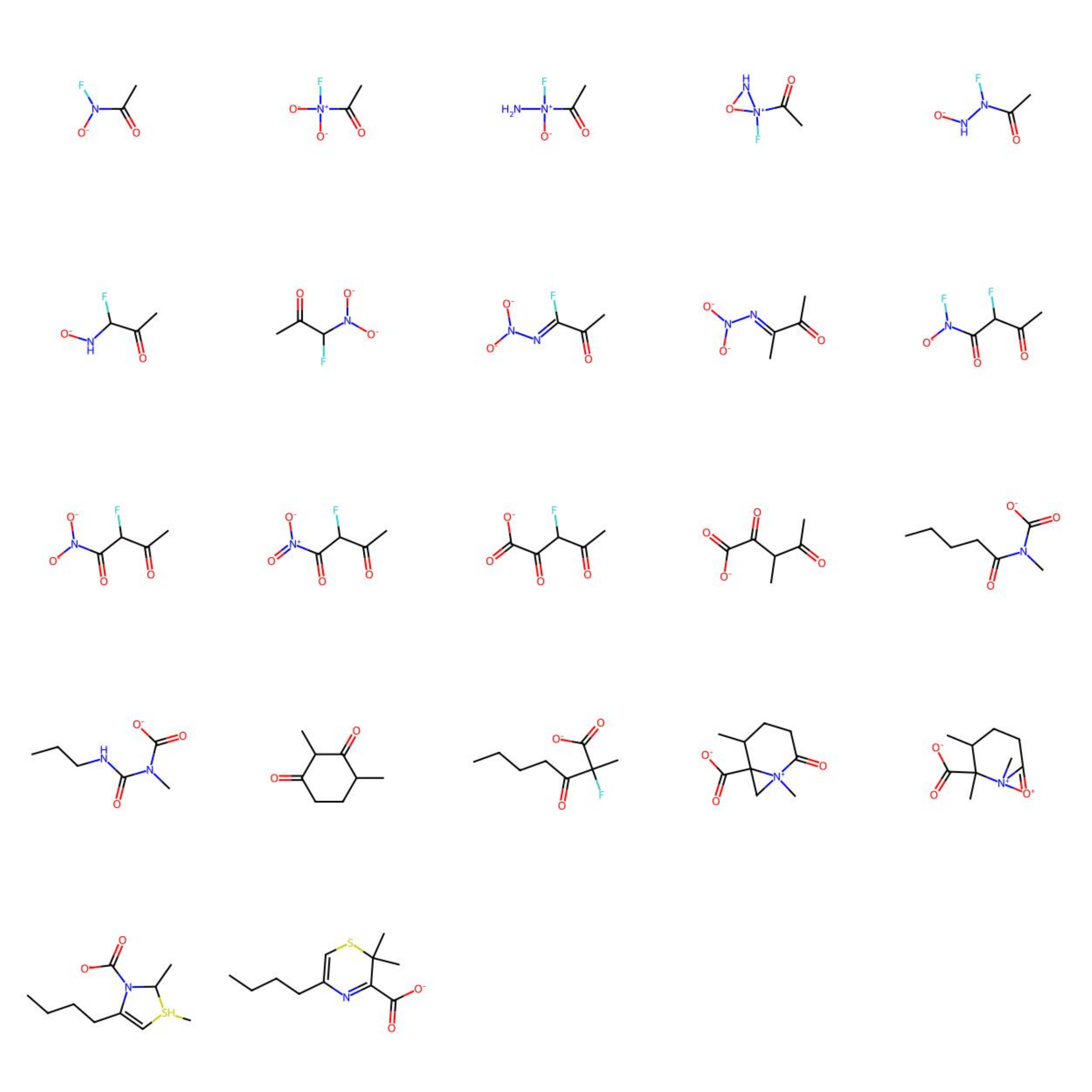}}
    \fbox{\includegraphics[width=0.9\linewidth]{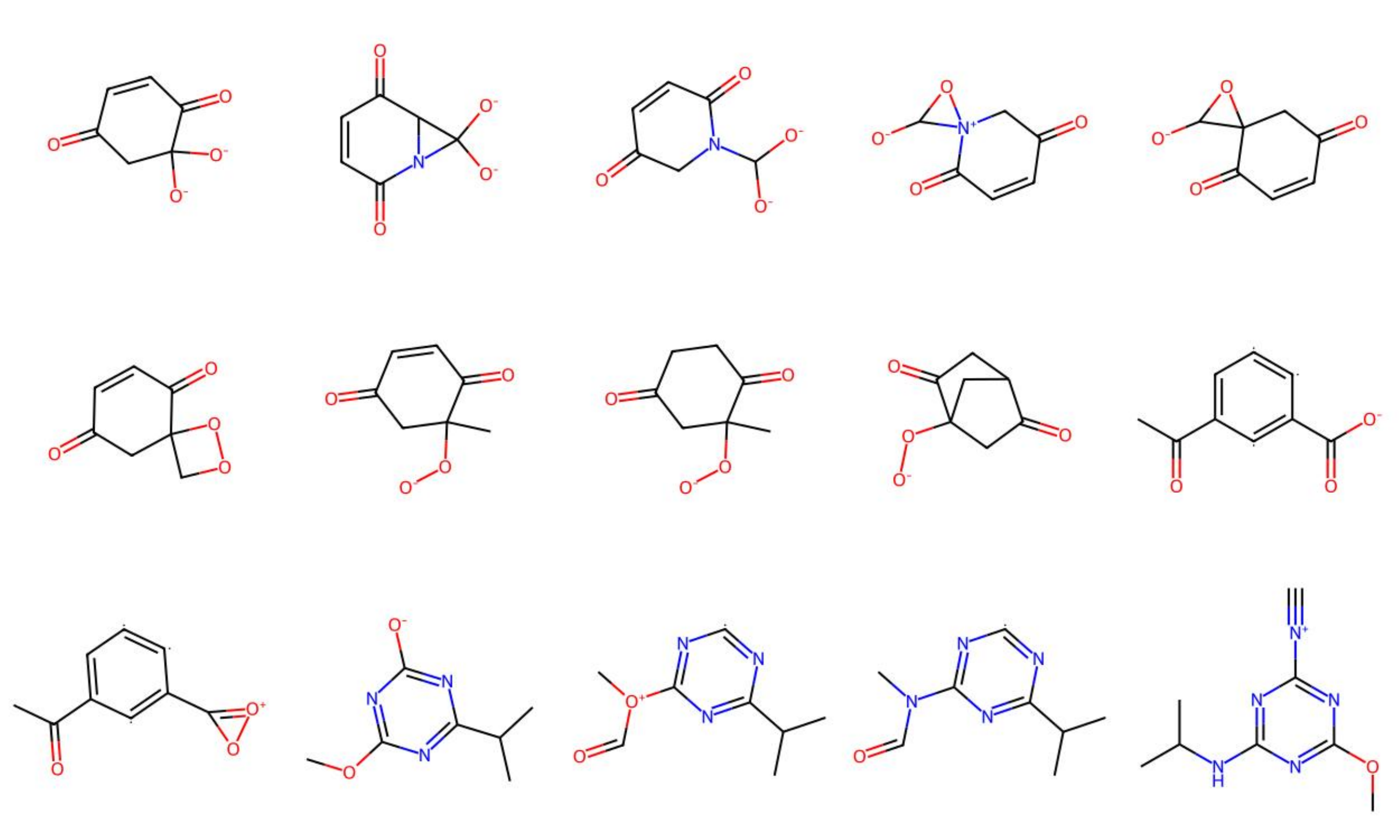}}
\end{figure}
\begin{figure}
    \centering
    \fbox{\includegraphics[width=0.9\linewidth]{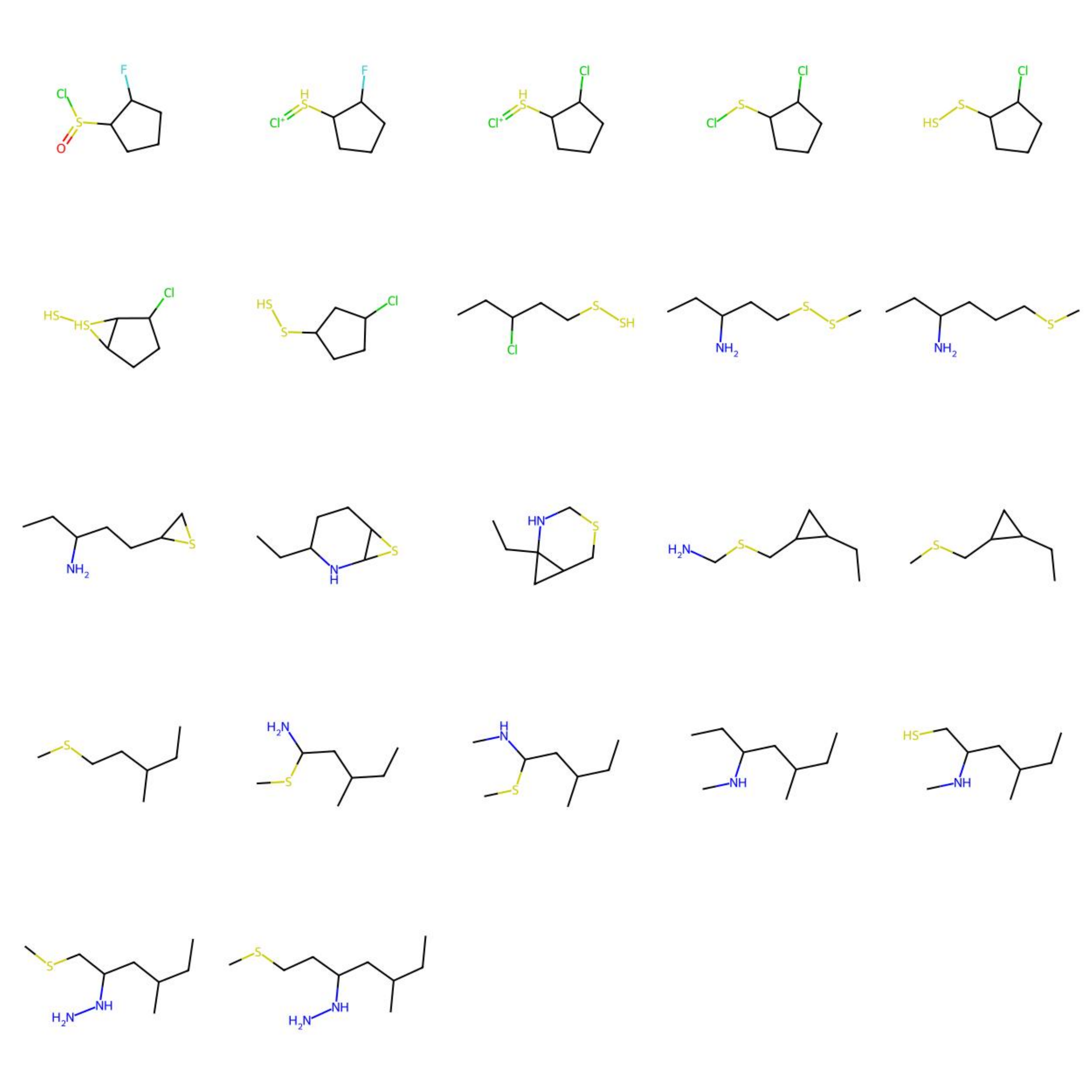}}
    \centering
    \fbox{\includegraphics[width=0.9\linewidth]{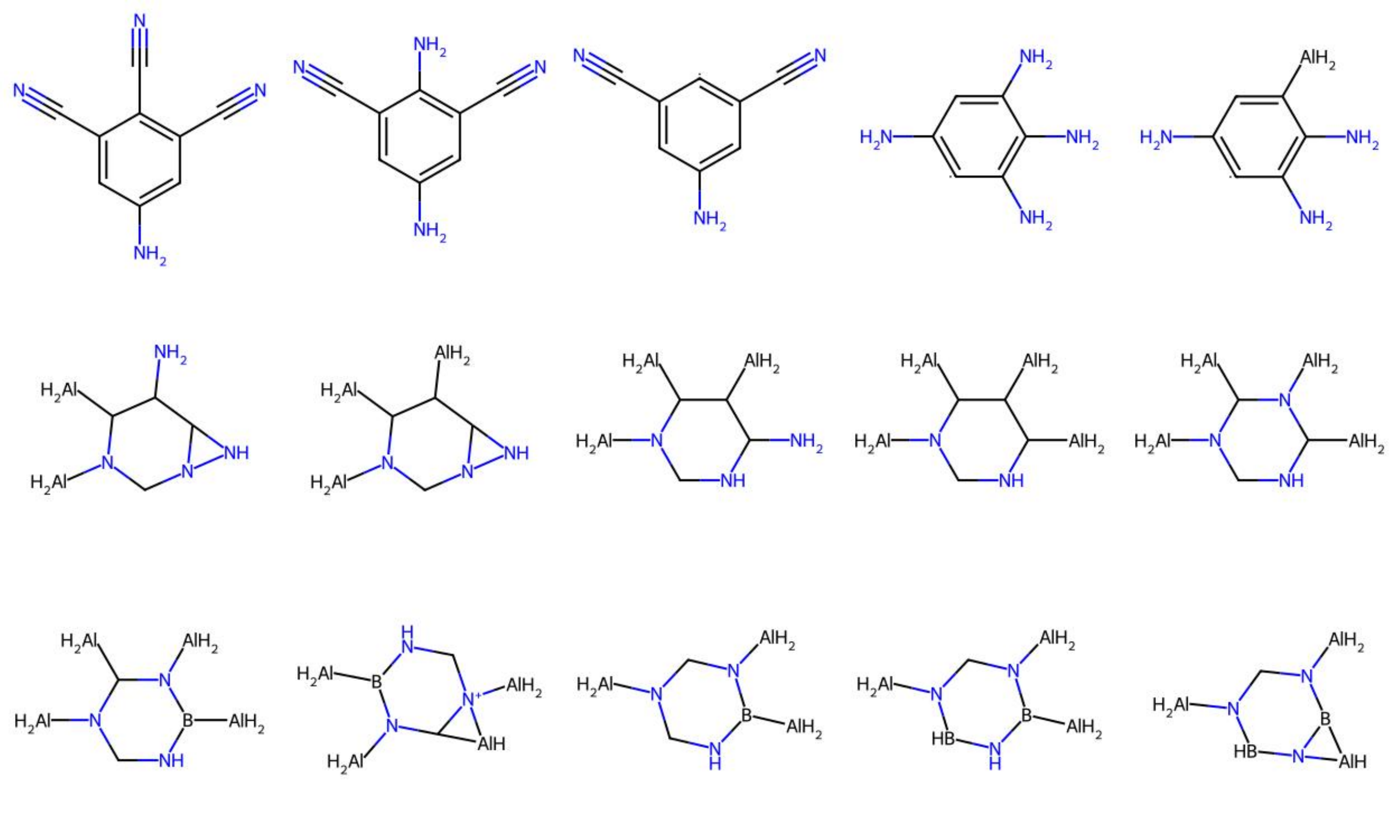}}
    \caption{Extra molecule interpolation plots (PC16)}
\end{figure}

\begin{figure}[H]
    \centering
    \fbox{\includegraphics[width=0.9\linewidth]{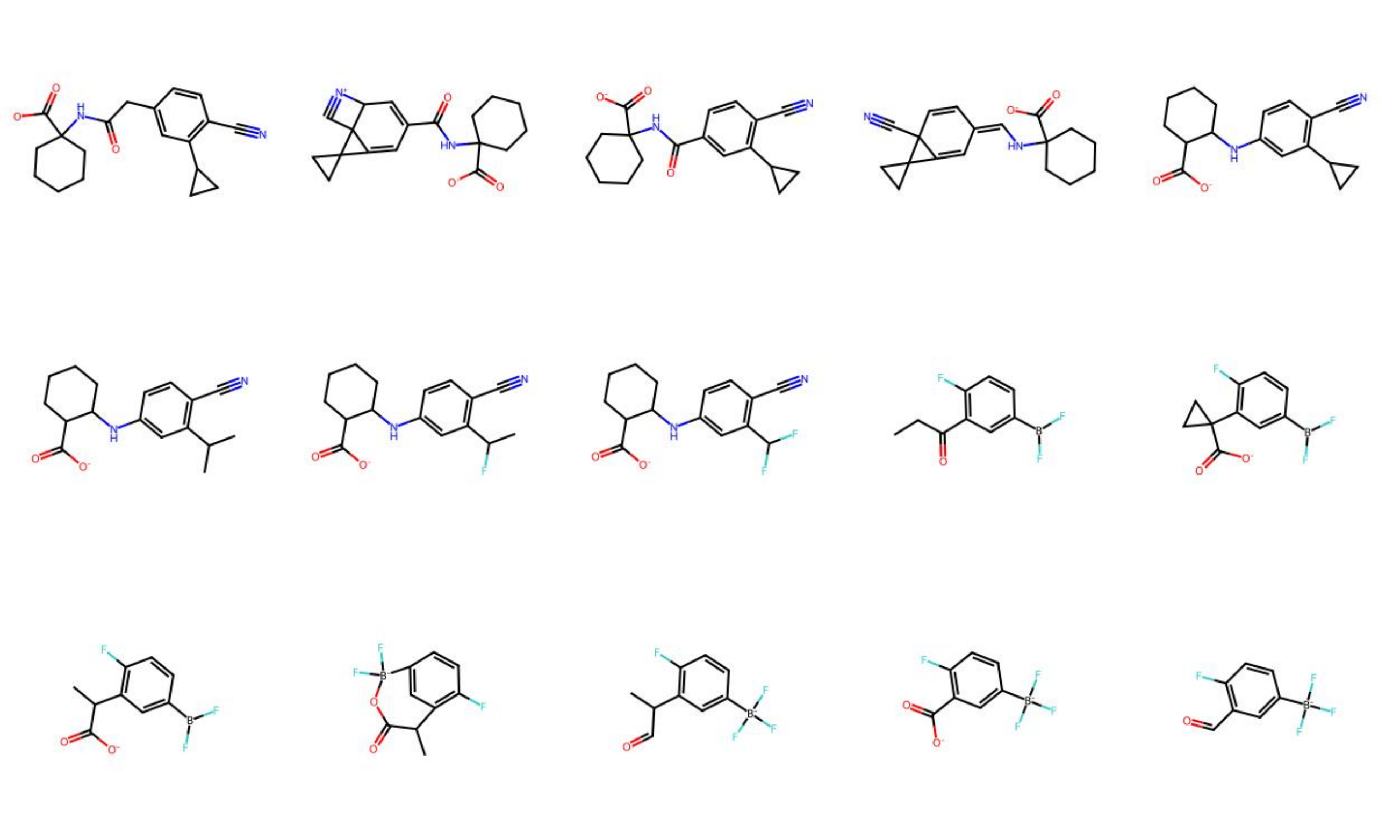}}
    \fbox{\includegraphics[width=0.9\linewidth]{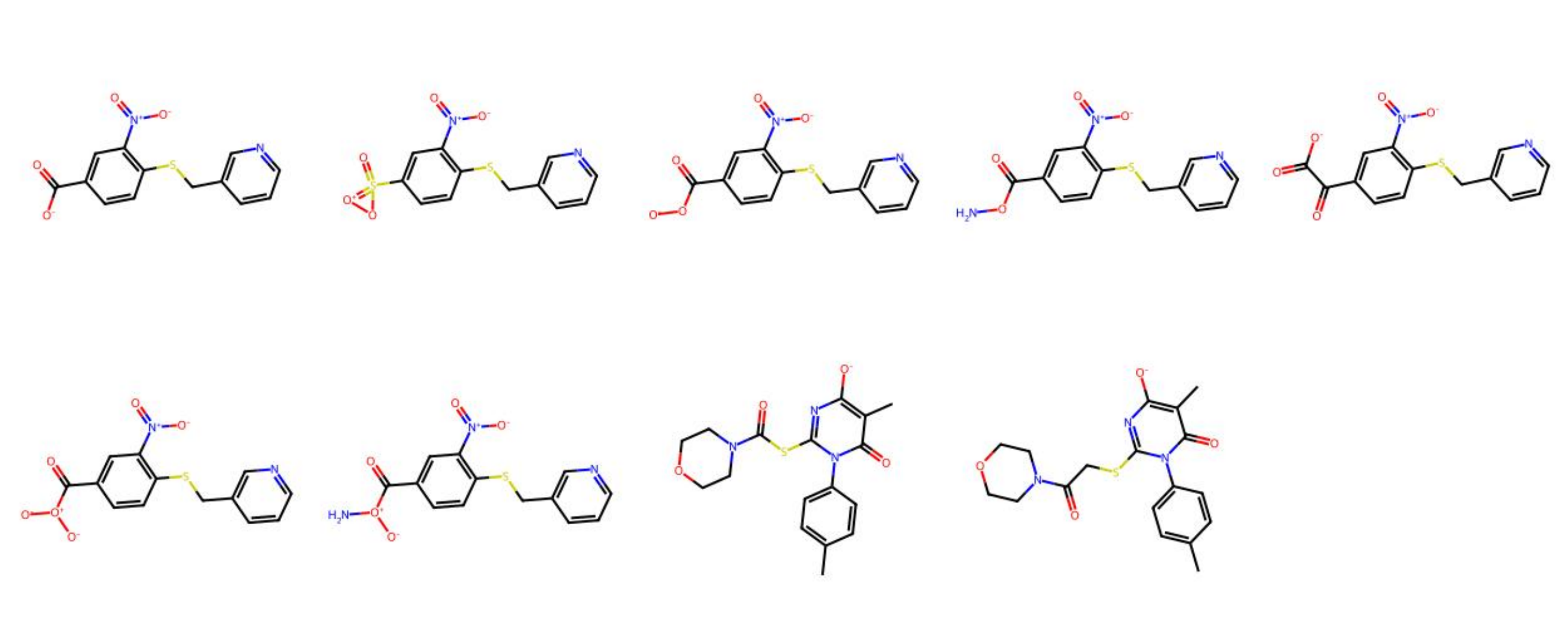}}
    \centering
    \fbox{\includegraphics[width=0.9\linewidth]{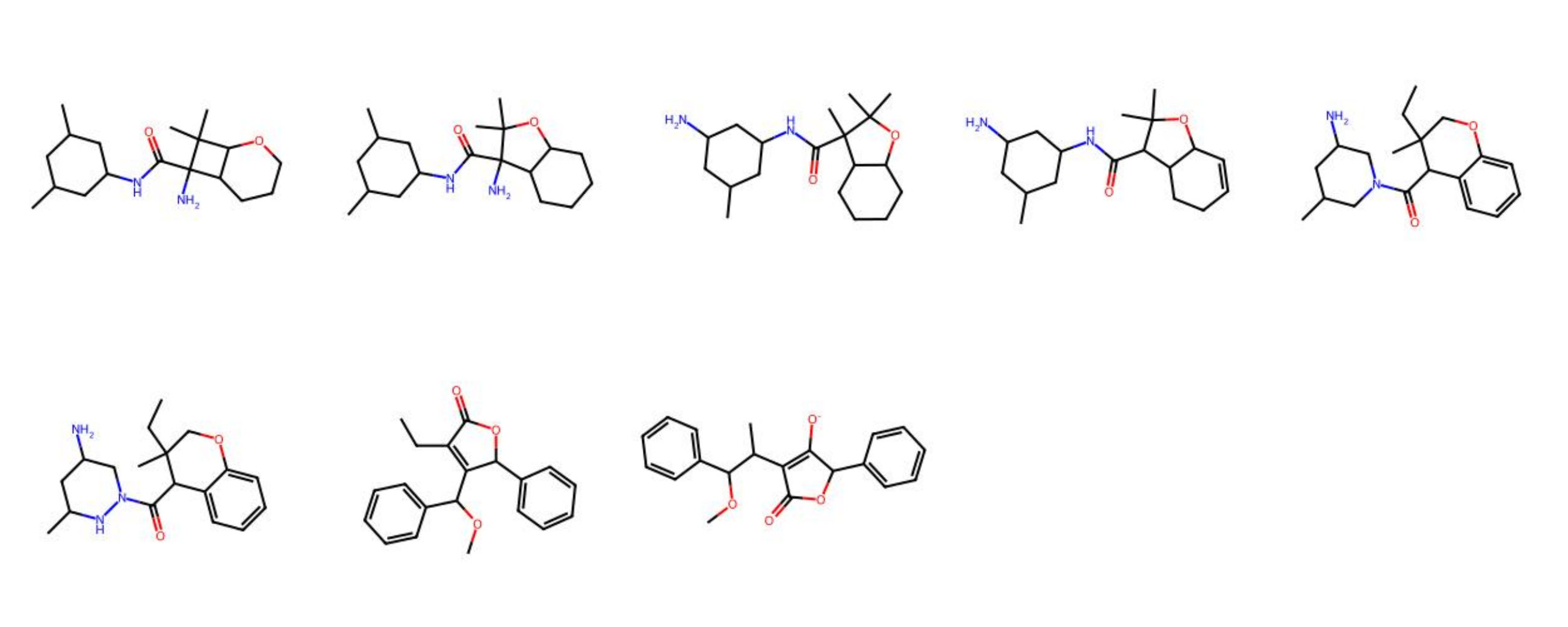}}
    \caption{Extra molecule interpolation plots (PC32)}
\end{figure}

\newpage
\section{Generated molecules samples}\label{app:gen_mols}

\begin{figure}[H]
    \centering
    \includegraphics[width=\linewidth]{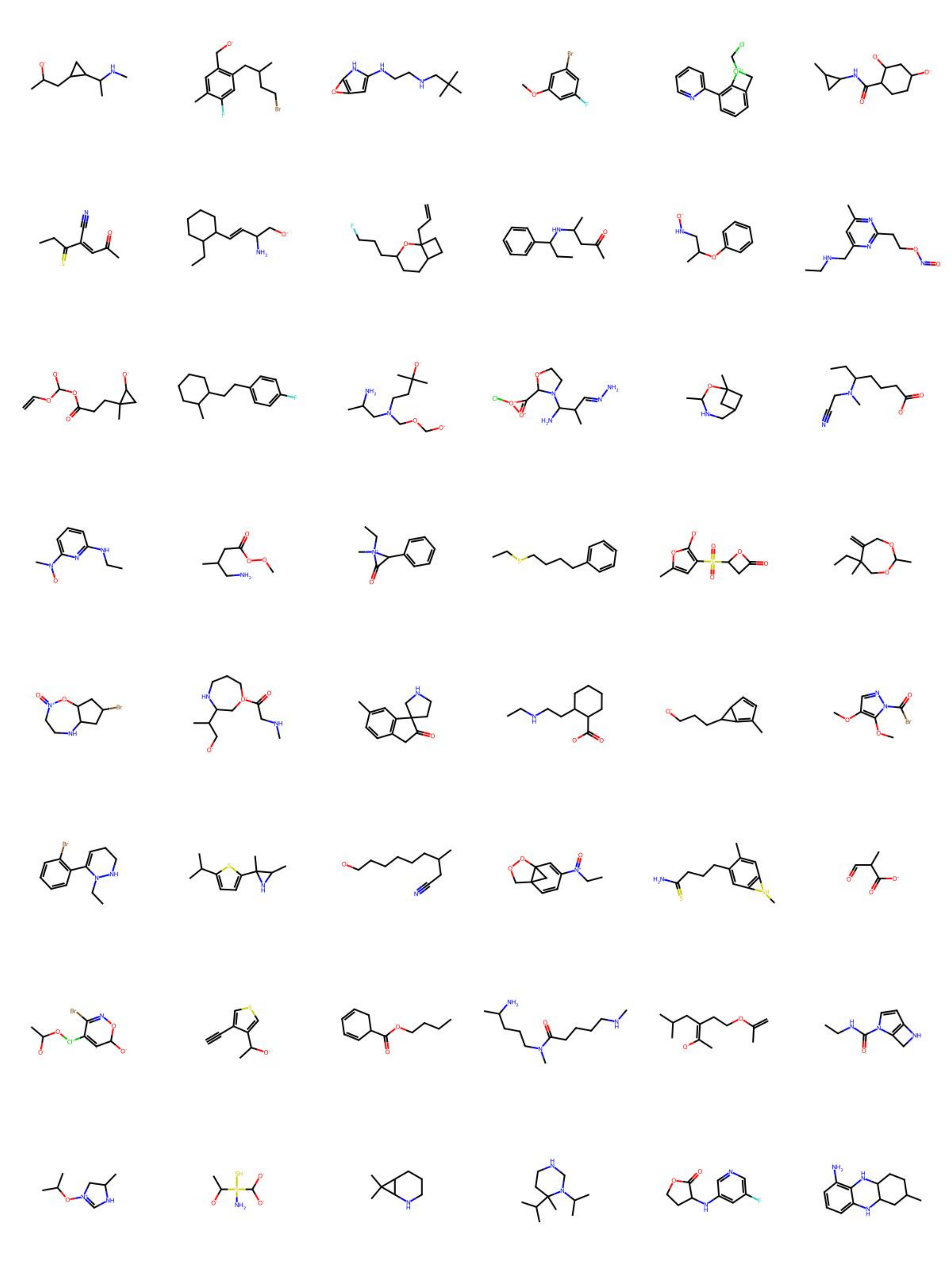}
    \caption{Samples of generated molecules (PubChem16)}
    \label{fig:gen_mol}
\end{figure}

Figure \ref{fig:gen_mol} presents examples of molecules generated by our PubChem‑trained model. Latent vectors are sampled from the standard normal prior and decoded into molecular graphs. We report only valid generations, that is, molecules for which all chemical constraints are satisfied.

\newpage

\section{Latent generation} \label{app:latentfm}

{ To further study the structure of our latent, we ran a simple flow matching algorithm to generate latents. The flow model is a simple MLP with FilM layers for time. Results are reported in table \ref{tab:gen_perf_fm}}.

\begin{table}[H]
    \caption{Generation performance of a flow matching trained to reconstruct latent representations of graphs. The model for each dataset is the variational autoencoder trained on kekulized molecules. We provide results for different numbers of generation steps. Statistics are computed on batches of 1024 molecules, 10 trials. Baselines are reported directly from \citep{roos2026categoricalflowmaps}.}
    \centering
    {\begin{tabular}{l c c c c c}
    \toprule
    \textsc{Dataset/Model} & \textsc{Steps} & \textsc{Valid. (\%)} & \textsc{Uniq. (\%)} & \textsc{Novelty (\%)} & \textsc{FCD}\\
    \midrule
    QM9/GraViti & VAE & 93.2 & 98.5 & N/A & 2.88\\
    \midrule
    QM9/GraViti & 1 & 94.1 & 99.2  & 
    N/A & 1.66 \\
    QM9/CFM ECLD & 1 & 95.8 & 89.2 & N/A & 2.16\\
    QM9/CFM CSD & 1 & 90.0 & 91.8 & N/A & 2.14\\
    \midrule
    QM9/GraViti & 2 & 95.6  & 99.8 & N/A & 0.75 \\
    QM9/CFM ECLD & 2 & 97.0 & 96.6 & N/A & 0.52\\
    QM9/CFM CSD & 2 & 91.1 & 97.6 & N/A & 0.49\\
    \midrule
    PubChem16/GraViti & VAE & 68.1  & 98.5 & 98.6  & 4.76 \\
    \midrule
    PubChem16/GraViti & 1 & 89.7 & 98.6 & 87.2 & 5.39\\
    PubChem16/GraViti & 2 & 85.9 & 100.0 & 97.1 & 2.74\\

    \bottomrule
    \end{tabular}}
    \label{tab:gen_perf_fm}
\end{table}

\end{document}